\documentclass[fleqn,10pt]{wlscirep}
\usepackage[utf8]{inputenc}
\usepackage[T1]{fontenc}
\usepackage{lineno}

\usepackage[utf8]{inputenc} 
\usepackage[T1]{fontenc}    
\usepackage{hyperref}       
\usepackage{url}            
\usepackage{booktabs}       
\usepackage{amsfonts}       
\usepackage{nicefrac}       
\usepackage{microtype}      
\usepackage{amsmath}
\usepackage{bbm}
\usepackage{graphicx}
\usepackage{subcaption}
\usepackage{wrapfig}
\usepackage{enumitem}
\usepackage{multirow}

\title{Molecular Contrastive Learning of Representations via Graph Neural Networks}

\author[1,2]{Yuyang Wang}
\author[3]{Jianren Wang}
\author[1]{Zhonglin Cao}
\author[1,2,4,*]{Amir Barati Farimani}
\affil[1]{Department of Mechanical Engineering, Carnegie Mellon University, Pittsburgh, PA 15213, USA}
\affil[2]{Machine Learning Department, Carnegie Mellon University, Pittsburgh, PA 15213, USA}
\affil[3]{Robotics Institute, Carnegie Mellon University, Pittsburgh, PA 15213, USA}
\affil[4]{Department of Chemical Engineering, Carnegie Mellon University, Pittsburgh, PA 15213, USA}
\affil[*]{corresponding author: Amir Barati Farimani (barati@cmu.edu)}

\begin{abstract}
Molecular Machine Learning (ML) bears promise for efficient molecule property prediction and drug discovery. However, labeled molecule data can be expensive and time-consuming to acquire. Due to the limited labeled data, it is a great challenge for supervised-learning ML models to generalize to the giant chemical space. In this work, we present \textit{MolCLR}: Molecular Contrastive Learning of Representations via Graph Neural Networks (GNNs), a self-supervised learning framework that leverages large unlabeled data ($\sim$10M unique molecules). In MolCLR pre-training, we build molecule graphs and develop GNN encoders to learn differentiable representations. Three molecule graph augmentations are proposed: atom masking, bond deletion, and subgraph removal. A contrastive estimator maximizes the agreement of augmentations from the same molecule while minimizing the agreement of different molecules. Experiments show that our contrastive learning framework significantly improves the performance of GNNs on various molecular property benchmarks including both classification and regression tasks. Benefiting from pre-training on the large unlabeled database, \textit{MolCLR} even achieves state-of-the-art on several challenging benchmarks after fine-tuning. Additionally, further investigations demonstrate that \textit{MolCLR} learns to embed molecules into representations that can distinguish chemically reasonable molecular similarities.
\end{abstract}

\begin{document}

\flushbottom
\maketitle




\section*{Introduction}

\begin{figure}[htb!]
    \centering
    \includegraphics[width=\textwidth, keepaspectratio=true]{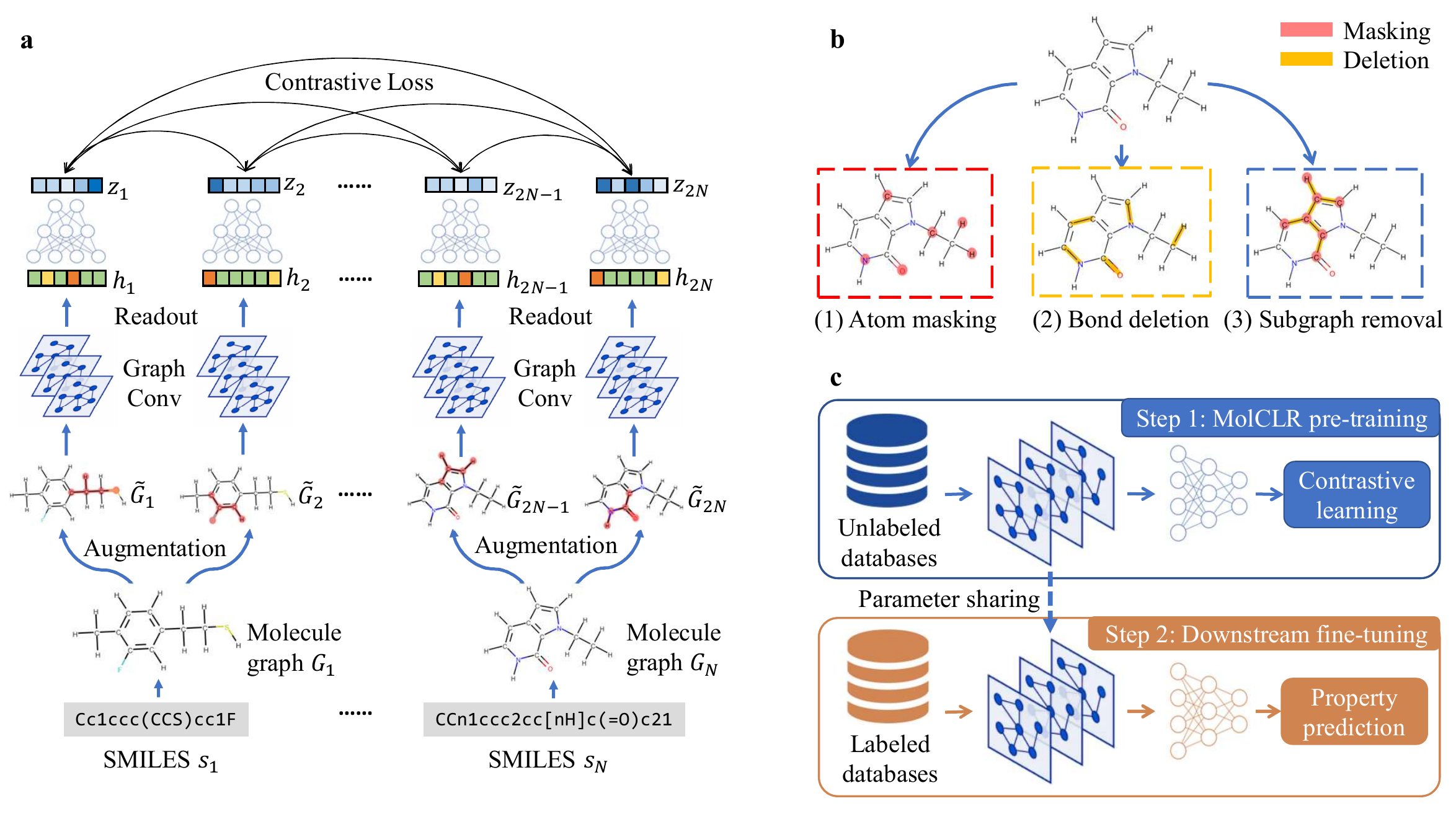}  \caption{Overview of MolCLR. (a) MolCLR pre-training: A SMILES $s_n$ from a mini-batch of $N$ molecule data is converted to a molecule graph $G_n$. Two stochastic molecule graph data augmentation operators are applied to each graph, resulting two correlated masked graphs: $\tilde{G}_{2n-1}$ and $\tilde{G}_{2n}$. A base feature encoder built upon graph convolutions and the readout operation extracts the representation $h_{2n-1}$, $h_{2n}$. Contrastive loss is utilized to maximize agreement between the latent vectors $z_{2n-1}$, $z_{2n}$ from the MLP projection head. (b) Molecule graph augmentation strategies: atom masking, bond deletion, and subgraph removal. (c) The whole MolCLR framework: GNNs are first pre-trained via MolCLR to learn representative features. Fine-tuning for downstream molecular property predictions shares the pre-trained parameters of the GNN encoder and randomly initializes an MLP head. It then follows the supervised learning to train the model. }
    \label{fig:pipeline}
\end{figure}

Molecular representation is fundamental and essential in the design of functional and novel chemical compounds \cite{bartok2013representing, huang2016communication, david2020molecular}. Due to the enormous magnitude of possible stable chemical compounds, development of an informative representation to generalize among the entire chemical space can be challenging \cite{oprea2001chemography}. Conventional molecular representations, like Extended-Connectivity Fingerprints (ECFP) \cite{rogers2010extended}, have became standard tools in computational chemistry. Recently, with the development of machine learning methods, data-driven molecular representation learning and its applications, including chemical property prediction \cite{duvenaud2015convfingerprint, gilmer2017neural, karamad2020orbital}, chemical modeling \cite{chmiela2018towards, deringer2018realistic, wang2019coarse}, and molecule design \cite{altae2017low, magar2021potential, wang2021efficient}, have gathered growing attentions. 

However, learning such representations can be difficult due to three major challenges. Firstly, it is hard to represent the molecular information thoroughly. For instance, string-based representations, like SMILES \cite{weininger1988smiles} and SELFIES \cite{krenn2020self}, fail to encode the important topology information directly. To preserve the rich structural information, many recent works exploit Graph Neural Networks (GNNs)~\cite{kipf2016semi, xu2018how}, and have shown promising results in molecular property prediction~\cite{gilmer2017neural, schutt2018schnet, yang2019analyzing}. Secondly, the magnitude of chemical space is enormous \cite{kirkpatrick2004chemical}, e.g., the size of potential pharmacologically active molecules is estimated to be in the order of $10^{60}$ \cite{bohacek1996art}. This places a great difficulty for any molecular representations to generalize among the potential chemical compounds. Thirdly, labeled data for molecular learning tasks are expensive and far from sufficient, especially when compared with the size of potential chemical space. Obtaining labels of molecular property usually requires sophisticated and time-consuming lab experiments \cite{brown2019guacamol}. The breadth of chemical research further complicates the challenges because the properties of interest range from quantum mechanics to biophysics \cite{wu2018moleculenet}. Consequently, the number of labels in most molecular learning benchmarks is far from adequate. Machine learning models trained on such limited data can easily get over-fit and perform poorly on molecules dissimilar to the training set. 

Molecular representation learning has been growing rapidly over the last decade with the development and success of machine learning, especially Deep Neural Networks (DNNs) \cite{lecun2015deep, duvenaud2015convfingerprint, vamathevan2019applications}. In conventional cheminformatics, molecules are represented in unique fingerprint (FP) vectors, such as ECFP. Given the FPs, DNNs are built to predict certain properties\cite{unterthiner2014deep,ma2015deep,ramsundar2015massively}. Besides the FP, string-based representations (e.g., SMILES) are widely used for molecular learning \cite{kusner2017grammar, gupta2018generative}. Language models built upon RNNs are a direct fit for learning representation from SMILES \cite{xu2017seq2seq, gomez2018automatic}. With the recent success of transformer-based architectures, such language models have been also utilized in learning representations from SMILES \cite{schwaller2019moleculartransformer, maziarka2020moleculeattention}. Recently, GNNs, which naturally encode the structure information, have been introduced to molecular representation learning \cite{duvenaud2015convfingerprint, feinberg2018potentialnet}. MPNN \cite{gilmer2017neural} and D-MPNN \cite{yang2019analyzing} implement a message-passing architecture to aggregate the information from molecule graphs. Further, SchNet \cite{schutt2018schnet} models quantum interactions within molecules in the GNN. DimeNet \cite{klicpera2020directional} integrates the directional information by transforming messages based on the angle between atoms.

Benefiting from the growth of available molecule data \cite{gaulton2012chembl,sterling2015zinc,wu2018moleculenet,kim2019pubchem}, self-supervised/pre-trained molecular representation learning has also been investigated. Self-supervised language models, like BERT \cite{devlin2018bert}, have been implemented to learn molecular representation with SMILES as input \cite{chithrananda2020chemberta, wang2019smiles}. On molecule graph, N-Gram Graph \cite{Liu2019NGramGS} builds the representation for the graph by assembling the vertex embedding in short walks, which needs no training. Hu et al. \cite{Hu2020Strategies} propose both node-level and graph-level tasks for GNN pre-training. However, the graph-level pre-training is based on supervised-learning tasks, which is constraint by limited labels. You et al. \cite{you2020graph} extends the contrastive learning to unstructured graph data, but the framework is not specifically designed for molecule graph learning and is only trained on limited molecular data. 

In this work, we propose MolCLR: Molecular Contrastive Learning of Representations via Graph Neural Networks shown in Figure~\ref{fig:pipeline} to address all the above challenges. MolCLR is a self-supervised learning framework trained on the large unlabeled molecular dataset with $\sim$10M unique molecules. Through contrastive loss \cite{oord2018infonce, chen2020simple}, MolCLR learns the representations by contrasting positive molecule graph pairs against negative ones. Three molecule graph augmentation strategies are introduced: atom masking, bond deletion, and subgraph removal. Molecule graph pairs augmented from the same molecule are denoted as positive, while others are denoted as negative. Widely-used GNN models, Graph Convolutional Network (GCN) \cite{kipf2016semi} and Graph Isomorphism Network (GIN) \cite{xu2018how}, are developed as GNN encoders in MolCLR to extract informative representation from molecule graphs. The pre-trained model is then fine-tuned on the downstream molecular property prediction benchmarks from MoleculeNet \cite{wu2018moleculenet}. In comparison to GCN and GIN trained via supervised learning, our MolCLR significantly improves the performance on both classification and regression tasks. Benefiting from pre-training on the large database, MolCLR surpasses other self-supervised learning and pre-training strategies in multiple molecular benchmarks. Moreover, on several tasks, our MolCLR rivals or even exceeds supervised learning baselines which include sophisticated graph convolution operations for molecules or domain-specific featurization. We also demonstrate that our molecule graph augmentation strategies improve the performance of supervised learning on molecular benchmarks when utilized as a direct data augmentation plug-in. Further comparison between MolCLR representations and conventional FPs indicates MolCLR learns to distinguish molecular similarities from pre-training on the large unlabeled data. 

To summarize, (1) We propose MolCLR, a self-supervised learning framework for molecular representation learning. (2) Three molecule graph augmentation strategies are introduced to generate contrastive pairs, namely atom masking, bond deletion, and subgraph removal. (3) Benefiting from pre-training on large unlabeled data, simple GNN models trained via MolCLR demonstrate significant improvements on all molecular benchmarks in comparison to supervised learning. (4) MolCLR even boosts simple GNN models to the state-of-the-art (SOTA) on several molecular benchmarks with fine-tuning, compared to more sophisticated GNNs which cannot utilize unlabeled data.

\section*{Results}

\subsection*{MolCLR Framework}

Our MolCLR model is developed upon the contrastive learning framework \cite{chen2020simple, chen2020big}. Latent representations from positive augmented molecule graph pairs are contrasted with representations from negative pairs. The whole pipeline (Figure~\ref{fig:pipeline}(a)) is composed of four components: data processing and augmentation, GNN-based feature extractor, non-linear projection head, and the normalized temperature-scaled cross-entropy (NT-Xent) \cite{chen2020simple} contrastive loss. 

Given a SMILES data $s_n$ from a mini-batch of size $N$, the corresponding molecule graph $G_n$ is built, in which each node represents an atom and each edge represents a chemical bond between atoms. Using molecule graph augmentation strategies, $G_n$ is transformed into two different but correlated molecule graphs: $\tilde{G}_i$ and $\tilde{G}_j$, where $i=2n-1$ and $j=2n$. Molecule graphs augmented from the same molecule are denoted as a positive pair, whereas those from different molecules are denoted as negative pairs. The feature extractor $f(\cdot)$ is modeled by GNNs and maps the molecule graphs into the representations $h_i, h_j \in \mathbb{R}^d$. In our case, we implement GCN \cite{kipf2016semi} and GIN \cite{xu2018how} with an average pooling as the feature extractor. A non-linear projection head $g(\cdot)$ is modeled by an MLP with one hidden layer, which maps the representations $h_i$ and $h_j$ into latent vectors $z_i$ and $z_j$, respectively. Normalized temperature cross-entropy (NT-Xent) loss \cite{chen2020simple} is applied to the $2N$ latent vectors $z$'s to maximize the agreement of positive pairs while minimizing the agreement of negative ones. The framework is pre-trained on the $\sim$10M unlabeled data from PubChem \cite{kim2019pubchem}.

The MolCLR pre-trained GNN models are fine-tuned for molecular property prediction as shown in Figure~\ref{fig:pipeline}(c). Similarly to the pre-training model, the prediction model consists of a GNN backbone and an MLP head, in which the former shares the same model as the pre-trained feature extractor, and the latter maps features into the predicted molecular property. The GNN backbone in the fine-tuning model is initialized by parameter sharing from the pre-trained model while the MLP head is initialized randomly. The whole fine-tuning model is then trained in a supervised learning manner on the target molecular property database. More details can be found in the Methods section. 

\subsection*{Molecule Graph Augmentation}
\label{sec:data_aug}

We employ three molecule graph data augmentation strategies (Figure~\ref{fig:pipeline}(b)) for input molecules in MolCLR: atom masking, bond deletion, and subgraph removal. 

\paragraph{Atom Masking} Atoms in the molecule graph are randomly masked with a given ratio. When an atom is masked, its atom feature $\pmb x_v$ is replaced by a mask token $\pmb m$, which is distinguished from any atom features in the molecule graph shown by the red box in Figure~\ref{fig:pipeline}(b). Through masking, the model is forced to learn the intrinsic chemical information (such as possible types of atoms connected by certain covalent bonds) within molecules. 

\paragraph{Bond Deletion} Bond deletion randomly deletes chemical bonds between the atoms with a certain ratio as the yellow box in Figure~\ref{fig:pipeline}(b) illustrates. Unlike atom masking which substitutes the original feature with a mask token, bond deletion is a more rigorous augmentation as it removes the edges completely from the molecule graph. Forming and breaking of chemical bonds between atoms determines the attributes of molecules in chemical reactions \cite{do2019graph}. Bond deletion mimics the breaking of chemical bonds which prompts the model to learn correlations between the involvements of one molecule in various reactions. 

\paragraph{Subgraph Removal} Subgraph removal can be considered as a combination of atom masking and bond deletion. Subgraph removal starts from a randomly picked origin atom. The removal process proceeds by masking the neighbors of the original atom, and then the neighbors of the neighbors, until the number of masked atoms reaches a given ratio of the total number of atoms. The bonds between the masked atoms are then deleted, such that the masked atoms and deleted bonds form an induced subgraph of the original molecule graph. As the blue box in Figure~\ref{fig:pipeline}(b) shows, the removed subgraph includes all the bonds between the masked atoms. By matching the molecule graphs with different substructures removed, the model learns to find the remarkable motifs within the remaining subgraphs \cite{jin2020hierarchical} which greatly determines the molecular properties.  

\subsection*{Molecular Property Predictions}

\begin{table}[htb!]
  \centering
  \begin{tabular}{llllllll}
    \toprule
    Dataset & BBBP & Tox21 & ClinTox & HIV & BACE & SIDER & MUV \\
    \# Molecules & 2039 & 7831 & 1478 & 41127 & 1513 & 1427 & 93087 \\
    \# Tasks & 1 & 12 & 2 & 1 & 1 & 27 & 17 \\
    \midrule
    RF & 71.4$\pm$0.0 & 76.9$\pm$1.5 & 71.3$\pm$5.6 & 78.1$\pm$0.6 & \textbf{86.7$\pm$0.8} & \textbf{68.4$\pm$0.9} & 63.2$\pm$2.3 \\
    SVM & 72.9$\pm$0.0 & \textbf{81.8$\pm$1.0} & 66.9$\pm$9.2 & \textbf{79.2$\pm$0.0}  & 86.2$\pm$0.0 & 68.2$\pm$1.3 & 67.3$\pm$1.3 \\
    GCN \cite{kipf2016semi} & 71.8$\pm$0.9 & 70.9$\pm$2.6 & 62.5$\pm$2.8 & 74.0$\pm$3.0 & 71.6$\pm$2.0 & 53.6$\pm$3.2 & 71.6$\pm$4.0 \\
    GIN \cite{xu2018how} & 65.8$\pm$4.5 & 74.0$\pm$0.8 & 58.0$\pm$4.4 & 75.3$\pm$1.9 & 70.1$\pm$5.4 & 57.3$\pm$1.6 & 71.8$\pm$2.5 \\
    SchNet \cite{schutt2018schnet} & 84.8$\pm$2.2 & 77.2$\pm$2.3 & 71.5$\pm$3.7 & 70.2$\pm$3.4 & 76.6$\pm$1.1 & 53.9$\pm$3.7 & 71.3$\pm$3.0 \\
    MGCN \cite{lu2019molecular} & \textbf{85.0$\pm$6.4} & 70.7$\pm$1.6 & 63.4$\pm$4.2 & 73.8$\pm$1.6  & 73.4$\pm$3.0 & 55.2$\pm$1.8 & 70.2$\pm$3.4 \\
    D-MPNN \cite{yang2019analyzing} & 71.2$\pm$3.8 & 68.9$\pm$1.3 & \textbf{90.5$\pm$5.3} & 75.0$\pm$2.1 & 85.3$\pm$5.3 & 63.2$\pm$2.3 & \textbf{76.2$\pm$2.8}  \\
    \midrule
    Hu et al. \cite{Hu2020Strategies} & 70.8$\pm$1.5 & 78.7$\pm$0.4 & 78.9$\pm$2.4 & 80.2$\pm$0.9 & 85.9$\pm$0.8 & 65.2$\pm$0.9 & 81.4$\pm$2.0 \\
    N-Gram \cite{Liu2019NGramGS} & \textbf{91.2$\pm$3.0} & 76.9$\pm$2.7 & 85.5$\pm$3.7 & \textbf{83.0$\pm$1.3} & 87.6$\pm$3.5 & 63.2$\pm$0.5 & 81.6$\pm$1.9 \\
    MolCLR\textsubscript{GCN} & 73.8$\pm$0.2 & 74.7$\pm$0.8 & 86.7$\pm$1.0 & 77.8$\pm$0.5 & 78.8$\pm$0.5 & 66.9$\pm$1.2 & 84.0$\pm$1.8 \\
    MolCLR\textsubscript{GIN} & 73.6$\pm$0.5 & \textbf{79.8$\pm$0.7} & \textbf{93.2$\pm$1.7} & 80.6$\pm$1.1 & \textbf{89.0$\pm$0.3} & \textbf{68.0$\pm$1.1} & \textbf{88.6$\pm$2.2} \\
    \bottomrule
  \end{tabular}
  \caption{Test performance of different models on seven classification benchmarks. The first seven models are supervised learning methods and the last four are self-supervised/pre-training methods. Mean and standard deviation of test ROC-AUC (\%) on each benchmark are reported.* \\
  \footnotesize{*Best performing supervised and self-supervised/pre-training methods for each benchmark are marked as bold.} }
  \label{tb:classification}
\end{table}

\begin{table}[htb!]
  \centering
  \begin{tabular}{lllllll}
    \toprule
    Dataset & FreeSolv & ESOL & Lipo & QM7 & QM8 & QM9 \\
    \# Molecules & 642 & 1128 & 4200 & 6830 & 21786 & 130829 \\
    \# Tasks & 1 & 1 & 1 & 1 & 12 & 8 \\
    \midrule
    RF & \textbf{2.03$\pm$0.22} & 1.07$\pm$0.19 & 0.88$\pm$0.04 & 122.7$\pm$4.2 & 0.0423$\pm$0.0021 & 16.061$\pm$0.019 \\
    SVM & 3.14$\pm$0.00 & 1.50$\pm$0.00 & 0.82$\pm$0.00 & 156.9$\pm$0.0 & 0.0543$\pm$0.0010 & 24.613$\pm$0.144 \\
    GCN \cite{kipf2016semi} & 2.87$\pm$0.14 & 1.43$\pm$0.05 & 0.85$\pm$0.08 & 122.9$\pm$2.2 & 0.0366$\pm$0.0011 & 5.796$\pm$1.969 \\
    GIN \cite{xu2018how} & 2.76$\pm$0.18 & 1.45$\pm$0.02 & 0.85$\pm$0.07 & 124.8$\pm$0.7 & 0.0371$\pm$0.0009 & 4.741$\pm$0.912 \\
    SchNet \cite{schutt2018schnet} & 3.22$\pm$0.76 & 1.05$\pm$0.06 & 0.91$\pm$0.10 & \textbf{74.2$\pm$6.0} & 0.0204$\pm$0.0021 & 0.081$\pm$0.001 \\
    MGCN \cite{lu2019molecular} & 3.35$\pm$0.01 & 1.27$\pm$0.15 & 1.11$\pm$0.04 & 77.6$\pm$4.7 & 0.0223$\pm$0.0021 & \textbf{0.050$\pm$0.002} \\
    D-MPNN \cite{yang2019analyzing} & 2.18$\pm$0.91 & \textbf{0.98$\pm$0.26} & \textbf{0.65$\pm$0.05} & 105.8$\pm$13.2 & \textbf{0.0143$\pm$0.0022} & 3.241$\pm$0.119 \\
    \midrule
    Hu et al. \cite{Hu2020Strategies} & 2.83$\pm$0.12 & 1.22$\pm$0.02 & 0.74$\pm$0.00 & 110.2$\pm$6.4 & 0.0191$\pm$0.0003 & 4.349$\pm$0.061 \\
    N-Gram \cite{Liu2019NGramGS} & 2.51$\pm$0.19 & \textbf{1.10$\pm$0.03} & 0.88$\pm$0.12 & 125.6$\pm$1.5 & 0.0320$\pm$0.0032 & 7.636$\pm$0.027 \\
    MolCLR\textsubscript{GCN} & 2.39$\pm$0.14 & 1.16$\pm$0.00 & 0.78$\pm$0.01 & \textbf{83.1$\pm$4.0} & 0.0181$\pm$0.0002 & 3.552$\pm$0.041 \\
    MolCLR\textsubscript{GIN} & \textbf{2.20$\pm$0.20} & \textbf{1.11$\pm$0.01} & \textbf{0.65$\pm$0.08} & 87.2$\pm$2.0 & \textbf{0.0174$\pm$0.0013} & \textbf{2.357$\pm$0.118} \\
    \bottomrule
  \end{tabular}
  \caption{Test performance of different models on six regression benchmarks. The first seven models are supervised learning methods and the last four are self-supervised/pre-training methods. Mean and standard deviation of test RMSE (for FreeSolv, ESOL, Lipo) or MAE (for QM7, QM8, QM9) are reported.* \\
  \footnotesize{*Best performing supervised and self-supervised/pre-training methods for each benchmark are marked as bold.}}
  \label{tb:regression}
\end{table}

To demonstrate the effectiveness of MolCLR, we benchmark the performance on multiple challenging classification and regression tasks from MoleculeNet \cite{wu2018moleculenet}. Details of molecular datasets can be found in Supplementary Table 1 and Supplementary Table 2. Table~\ref{tb:classification} shows the test ROC-AUC (\%) of our MolCLR model on classification tasks in comparison to supervised and self-supervised/pre-training baseline models. The average and standard deviation of three individual runs are reported. MolCLR\textsubscript{GCN} and MolCLR\textsubscript{GIN} denotes MolCLR pre-training with GCN and GIN as feature extractors, respectively. Observations from Table~\ref{tb:classification} are the followings. (1) In comparison with other self-supervised learning or pre-training strategies, our MolCLR framework achieves the best performance on 5 out of 7 benchmarks, with an average improvement of 4.0\%. Such improvement illustrates that our MolCLR is a powerful self-supervised learning strategy, which is easy to implement and requires little domain-specific sophistication. (2) Compared with best-performing supervised learning baselines, MolCLR also shows rival performance. In some benchmarks (e.g., ClinTox, BACE, MUV), our pre-training model even surpasses the SOTA supervised learning methods, which include sophisticated aggregation operations or domain-specific featurization. For instance, on ClinTox, MolCLR improves the ROC-AUC by 2.7\% with respect to supervised D-MPNN. (3) Notably, MolCLR performs remarkably well on datasets with a limited number of molecules, like ClinTox, BACE, and SIDER. The performance validates that MolCLR learns informative representations that can be transferred among different datasets. 

Table~\ref{tb:regression} includes the test performance of MolCLR and baseline models on regression benchmarks. FreeSolv, ESOL, and Lipo use root-mean-square error (RMSE) as the evaluation metric while QM7, QM8, and QM9 are measured via mean-absolute error (MAE), following the recommendation from MoleculeNet \cite{wu2018moleculenet}. Regression tasks are more challenging in comparison with classification since the latter only considers manually-defined discrete labels. Observations from Table~\ref{tb:regression} are the followings. (1) MolCLR surpasses other pre-training baselines in 5 out of 6 benchmarks and achieves almost the same performance on the remaining ESOL benchmark. Compared to Hu et al. \cite{Hu2020Strategies}, which also implements GIN as the encoder, MolCLR\textsubscript{GIN} outperforms it on all the 6 regression databases. On QM7 and QM9, for example, the improvement ratios over Hu et al. are 20.9\% and 45.8\%, respectively. (2) In comparison with supervised learning models, MolCLR reaches competitive performance in most cases. For example, MolCLR obtains similar results as the best performing supervised D-MPNN on Lipo database. Also, GCN and GIN achieve better prediction performance via MolCLR pre-training on all regression benchmarks. Although, in QM9, MolCLR does not rival with supervised SchNet \cite{schutt2018schnet} and MGCN \cite{lu2019molecular}. As the two models are specifically designed for quantum interaction and make use of extra 3D positional information. Notably, though SchNet and MGCN demonstrate superior performance on datasets concerning quantum mechanics properties (i.e., QM7, QM8, and QM9), while they do not show advantages over other supervised learning baselines on remaining benchmarks. Moreover, MolCLR pre-training is still demonstrated to be effective on the challenging QM9 benchmark. In comparison to GCN and GIN without pre-training, MolCLR still greatly boosts the performance by 38.7\% and 50.3\%, respectively. Also, MolCLR performs better than other self-supervised learning baselines on QM9, which validates the efficacy of MolCLR. Since properties in QM9 have of various units and magnitudes, detailed results of QM9 are reported in Supplementary Table 3. 

Both Table~\ref{tb:classification} and Table~\ref{tb:regression} show MolCLR pre-training greatly improves the performance on all the benchmarks compared to supervised GCN and GIN, which demonstrates the effectiveness of MolCLR. On classification benchmarks, the average gains via MolCLR are 12.4\% for GCN and 16.8\% for GIN. Similarly, on regressions, the averaged improvement ratios are 27.6\% for GCN and 33.5\% for GCN. In general, GIN demonstrates more improvement than GCN through MolCLR pre-training. This could be because GIN has more parameters and are capable of learning more representative molecular features. Also, MolCLR shows better prediction accuracy in comparison to other pre-training/self-supervised learning baselines in most cases. It should be emphasized that MolCLR benefits from pre-training on large unlabeled databases while the other supervised/self-supervised learning baselines do not. Leverage of unlabeled data provides a great advantage for MolCLR over other baselines in generalizing among the chemical space and various molecular properties. Influence of the pre-training database on MolCLR is further investigated in Supplementary Table 4 and Supplementary Figure 1. Such capability of generalization bears promises for predicting potential molecular properties in drug discovery and design. 

\subsection*{Optimal Molecule Graph Augmentations}

\begin{figure}
    \centering
    \includegraphics[width=\textwidth, keepaspectratio=true]{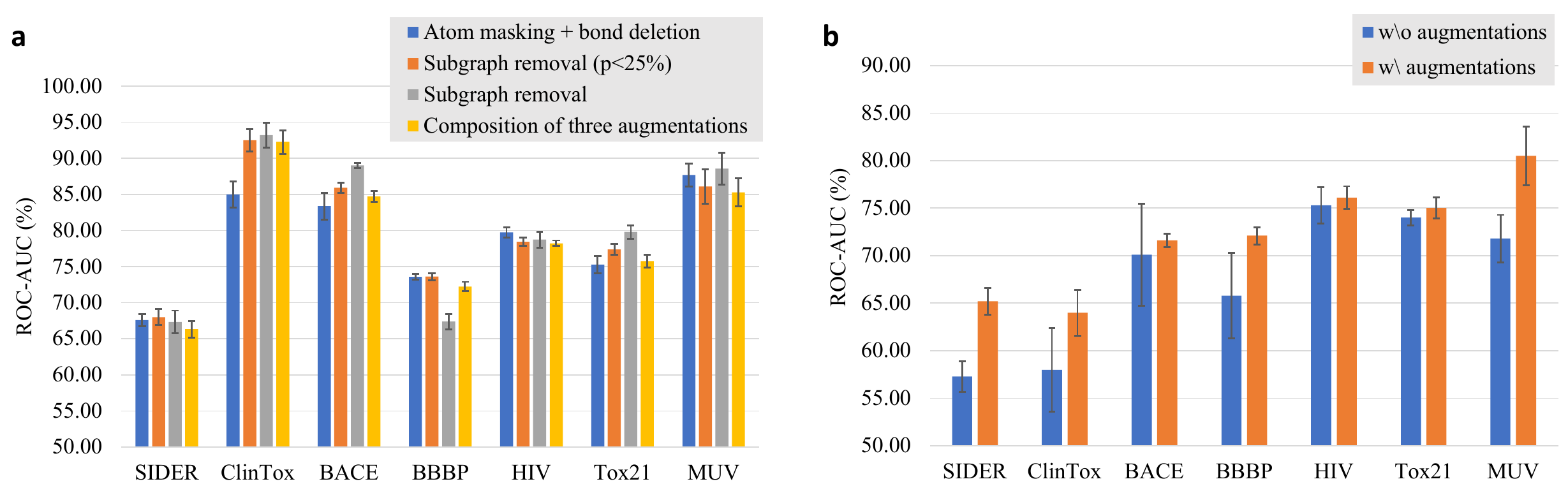}
\caption{Investigation of molecule graph augmentations on classification benchmarks. (a) Test performance of MolCLR models with different compositions of molecule graph augmentation strategies. (b) Test performance of GIN models trained via supervised learning with and without molecular graph augmentations. The height of each bar represents the mean ROC-AUC (\%) on the benchmark, and the length of each error bar represents the standard deviation.}
\label{fig:data_aug}
\end{figure}

To systematically analyze the effect of molecule graph augmentation strategies, we compare different compositions of atom masking, bond deletion, and subgraph removal. Shown in Figure~\ref{fig:data_aug}(a) are the ROC-AUC (\%) mean and standard deviation of each data augmentation strategy on different benchmarks. Four augmentation strategies are considered. (1) Integration of atom masking and bond deletion with both ratios $p$ set to 25\%. (2) Subgraph removal with a random ratio $p$ from 0\% to 25\%. (3) Subgraph removal with a fixed 25\% ratio. (4) Composition of all the three augmentation methods. Specifically, a subgraph removal with a random ratio 0\% to 25\% is applied at first. Then if the ratio of masked atoms is smaller than 25\%, we continue to randomly mask atoms until it reaches the ratio of 25\%. Similarly, if the bond deletion ratio is smaller than 25\%, more bonds are deleted to reach the set ratio. 

As Figure~\ref{fig:data_aug}(a) illustrates, subgraph removal with a 25\% ratio reaches the best performance on average among all the four compositions. Its outstanding performance can be attributed to that subgraph removal is an intrinsic combination of atom masking and bond deletion, and that subgraph removal further disentangles the local substructures compared with strategy (1). However, subgraph removal with a fixed 25\% ratio performs poorly in BBBP dataset because the molecule structures in BBBP are sensitive, such that a slight topology change can cause great property difference. Besides, it is worth noticing that the composition of all three augmentations (strategy (4)) hurts the ROC-AUC compared with single subgraph removal augmentation in most benchmarks. A possible reason is that the composition of all the three augmentation strategies can remove a wide range of substructures within the molecule graph, thus eliminate the important topology information. In general, subgraph removal achieves superior performance in most benchmarks. However, it is also indicated that the optimal molecule graph augmentation is task-independent. 

\subsection*{Molecule Graph Augmentation on Supervised Learning}

The molecule graph augmentation strategies in our work, namely atom masking, bond deletion, and subgraph removal, can be implemented as a generic data augmentation plug-in for any graph-based molecular learning methods. To validate the effectiveness of molecule graph augmentations on supervised molecular tasks, we train GIN models with and without augmentations from random initialization. Specifically, subgraph masking with a fixed ratio 25\% is implemented as the augmentation. Figure~\ref{fig:data_aug}(b) documents the mean and standard deviation of test ROC-AUC (\%) over the seven molecular property classification benchmarks. On all the seven benchmarks, GINs trained with augmentations surpass the models without augmentations. Molecule graph augmentations improve the averaged ROC-AUC score by 7.2\%. Implementation of our molecule graph augmentation strategies on supervised molecular property prediction tasks improves the performance greatly even without pre-training. It is indicated that molecule graph augmentations are effective in helping GNNs learn robust and representative features. For instance, subgraph removal matches partially observed molecule graphs. Therefore, the model learns to find the remarkable motifs within the remaining subgraphs which greatly benefits molecular property learning.

\subsection*{Investigation of MolCLR Representation}

\begin{figure}[htb!]
    \centering
    \includegraphics[width=0.9\textwidth, keepaspectratio=true]{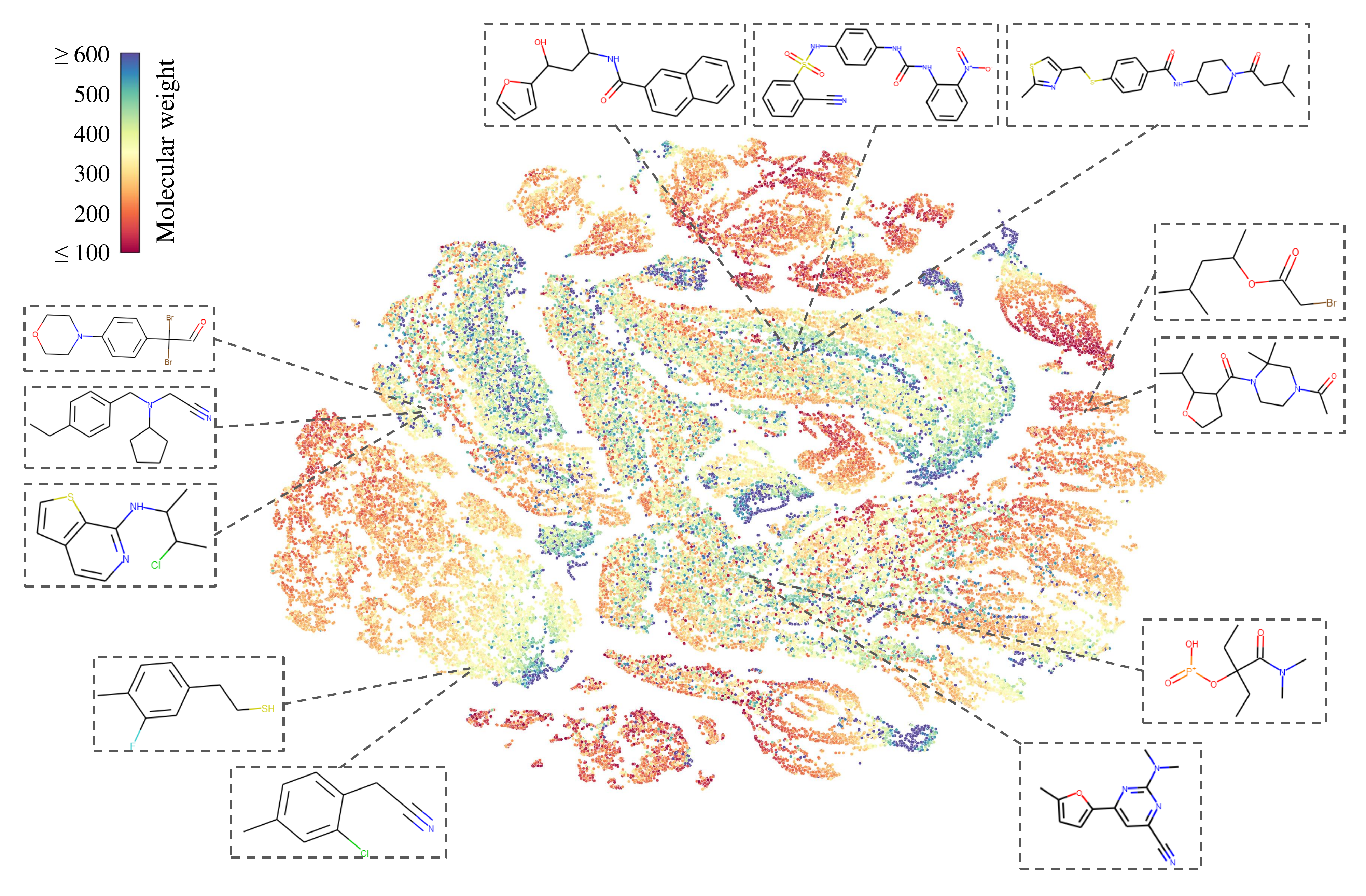}
    \caption{Visualization of molecular representations learned by MolCLR via t-SNE. Representations are extracted from the validation set of the pre-training dataset, which contains 100k unique molecules. Each point is colored by its corresponding molecular weight. Some molecules close in the representation domain are also shown. }
    \label{fig:tsne}
\end{figure}

We examine the representations learned by pre-trained MolCLR using t-SNE embedding \cite{van2008visualizing}. The t-SNE algorithm maps similar molecular representations to adjacent points in two-dimension (2D). Shown in Figure~\ref{fig:tsne} are 100K molecules from the validation set of the PubChem database embedded to 2D via t-SNE, colored based on the molecular weights. We also include some randomly selected molecules in the figure to illustrate what are the similar/dissimilar molecules learned by MolCLR pre-training. As shown in Figure~\ref{fig:tsne}, MolCLR learns close representations for molecules with similar topology structures and functional groups. For instance, the three molecules shown on the top possess carbonyl groups connected with aryls. The two molecules shown on the bottom left have similar structures, where a halogen atom (Fluorine or Chlorine) is connected to benzene. This illustrates that even without labels, the model learns intrinsic connections between molecules as molecules with similar properties have close features. More visualizations of MolCLR representations can be found in Supplementary Figure 2.

\begin{figure}[htb!]
    \centering
    \includegraphics[width=\textwidth, keepaspectratio=true]{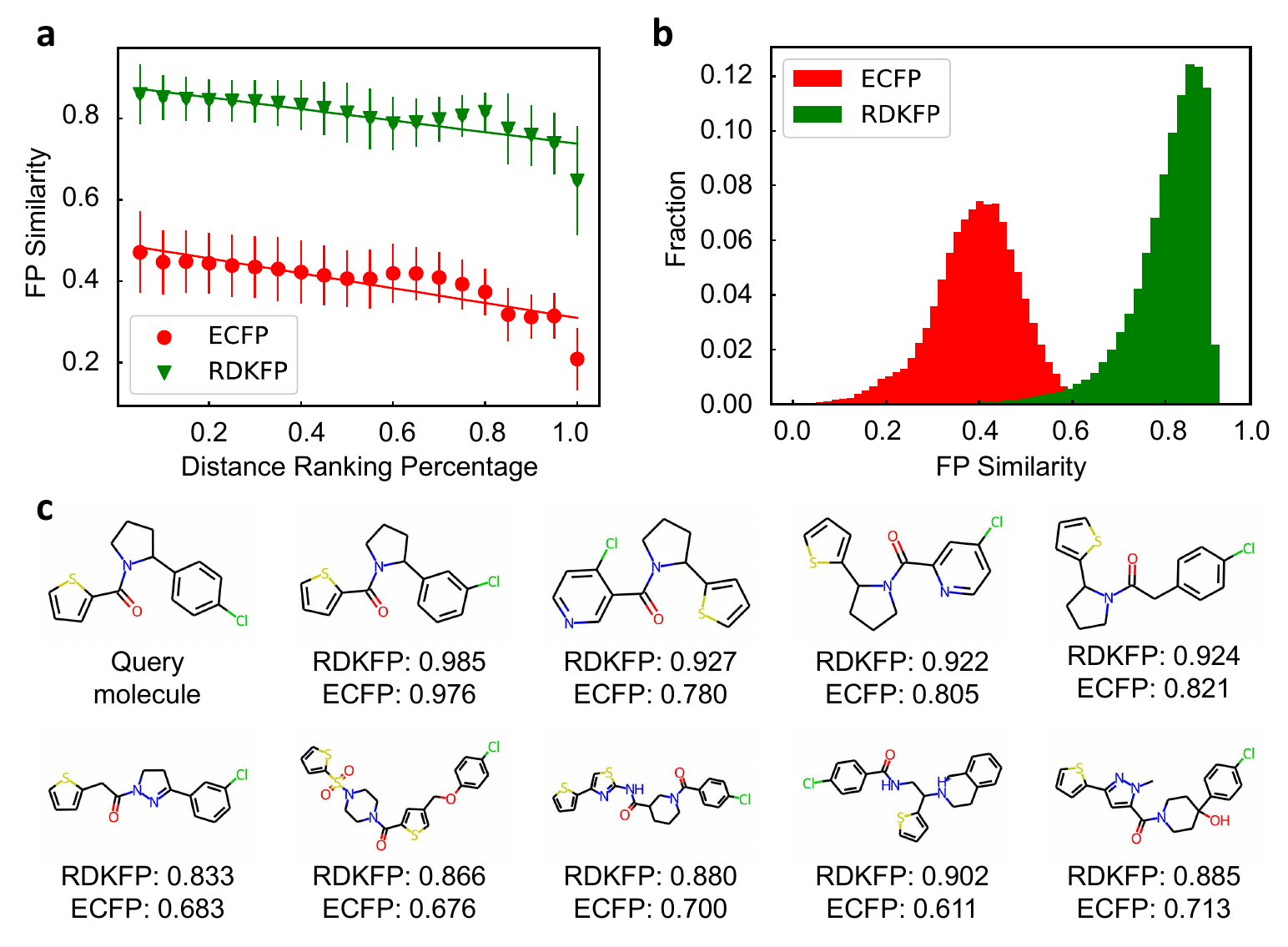}
\caption{Comparison of MolCLR-learned representations and conventional FPs using the query molecule (PubChem ID 42953211). (a) Change of ECFP and RDKFP similarities with respect to the distance between MolCLR representations. (b) Distribution of ECFP and EDKFP similarities with the query molecule. (c) The query molecule and 9 closest molecules in MolCLR representation domain with RDKFP and ECFP similarities labeled.}
\label{fig:sim}
\end{figure}

To further evaluate MolCLR, we compare the MolCLR-learned representations with conventional molecular fingerprints (FPs), e.g., ECFP \cite{rogers2010extended} and RDKFP. In particular, given a query molecule, we extract its representation via MolCLR and calculate its cosine distances with all reference molecules in our pre-training database. Cosine distance between two representations $(\pmb{u}, \pmb{v})$ are defines as $1-\frac{\pmb{u} \cdot \pmb{v}}{\|\pmb{u}\|\|\pmb{v}\|}$. All reference molecules are then ranked by the representation distances and uniformly divided into 20 bins based on the ranking percentage. The lower the percentage threshold is, the more similar molecules are expected with respect to the query, as the MolCLR representations are closer. Within each bin, 5000 molecules are randomly selected and their dice FP similarities with the query are calculated. Figure~\ref{fig:sim} shows an example of a query molecule (PubChem ID 42953211). Shown in Figure~\ref{fig:sim}(a) are the mean and standard deviation of FP similarities within each bin. The distribution of similarities using both ECFP and RDKFP are shown in Figure~\ref{fig:sim}(b). ECFP tends to obtain lower similarities than RDKFP since the former covers a wider range of features relevant to molecular activity. It is shown, though, as the MolCLR representation distance increases, both the ECFP and RDKFP similarities decrease. The averaged RDKFP similarities at the top 5\% is $\sim$0.9 and drops to $\sim$0.67 at the last 5\%. Similarly, the averaged ECFP similarity drops from $\sim$0.49 at the top 5\% to $\sim$0.21 at the last 5\%. Though there are fluctuations as the percentage threshold increases, the overall tendencies are consistent among the MolCLR learned representations and chemical FPs. Namely, the distance between MolCLR representations effectively reflects the molecular similarity. Besides, 9 molecules that are closest to the query molecule in the MolCLR representation domain are illustrated in Figure~\ref{fig:sim}(c) with both FPs similarities labeled. These molecules share high RDKFP similarities from 0.833 to 0.985, which further demonstrate MolCLR learns chemically meaningful representations. It is observed that these selected molecules share the same functional groups, including alkyl halides (chlorine), tertiary amines, ketones, and aromatics. A thiophene structure can also be found in all the molecules. Notably, the second molecule in the first row in Figure~\ref{fig:sim}(c) is exactly the same as the query molecule except for the position of the chlorine, hence the highest similarities. It is indicated that through contrastive learning on large unlabeled data, MolCLR automatically embeds molecules to representative features and distinguishes the compounds in a chemically reasonable manner. More examples of query molecules can be found in Supplementary Figure 3.

\section*{Conclusion}

In this work, we investigate self-supervised learning for molecular representation. Specifically, we propose Molecular Contrastive Learning of Representations (MolCLR) via GNNs and three molecular graph augmentations strategies: atom masking, bond deletion, and subgraph removal. Through contrasting positive pairs against negative pairs from augmentations, MolCLR learns informative representation with general GNN backbones. Experiments show that MolCLR pre-trained GNN models achieve great improvement on various molecular benchmarks, and show better generalizations compared with models trained in the supervised learning manner. 

Molecular representations learned by MolCLR demonstrate the transferability to molecular tasks with limited data and the power of generalization on the large chemical space. There are many promising directions to investigate as future works. For instance, improvement of the GNN backbones (e.g. transformer-based GNN architectures \cite{yun2019GraphTransformer}) can help extract better molecular representations. Besides, visualization and interpretation of self-supervised learned representations are of great interest \cite{pope2019explainability}. Such investigations can help researchers better understand chemical compounds and benefit drug discovery.

\section*{Methods}

\subsection*{Graph Neural Networks}

In our work, a molecule graph $G$ is defined as $G=(V,E)$, where $V$ and $E$ are nodes (atoms) and edges (chemical bonds), respectively \cite{bronstein2017geometric}. Modern Graph Neural Networks (GNNs) utilize a neighborhood aggregation operation, which updates the node representation iteratively \cite{kipf2016semi}. The aggregation update rule for a node feature on the $k$-th layer of a GNN is given in Equation~\ref{eq:aggregate}:
\begin{equation}
    \pmb{a}_v^{(k)} = \text{AGGREGATE}^{(k)}(\{\pmb h_u^{(k-1)}: u \in \mathcal{N}(v)\}), \;  \pmb h_v^{(k)} = \text{COMBINE}^{(k)}(\pmb h_v^{(k-1)}, \pmb a_v^{(k)}), 
    \label{eq:aggregate}
\end{equation}
where $\pmb h_v^{(k)}$ is the feature of node $v$ at the $k$-th layer and $\pmb h_v^{(0)}$ is initialized by node feature $\pmb x_v$. $\mathcal{N}(v)$ denotes the set of all the neighbors of node $v$. To further extract a graph-level feature $\pmb h_G$, readout operation integrates all the node features among the graph $G$ as given in Equation~\ref{eq:readout}:
\begin{equation}
    \pmb h_G = \text{READOUT}(\{\pmb h_u^{(k)}: v \in G\}).
    \label{eq:readout}
\end{equation}
In our work, we build GNN encoders based on GCN \cite{kipf2016semi} and GIN \cite{xu2018how}. GCN integrates the aggregation and combination operations by introducing a mean-pooling over the node itself and its adjacencies before the linear transformation. While GIN utilizes an MLP and weighted summation of node features in the aggregation. Both are simple yet generic graph convolutional operations. Additionally, we implement widely-used mean pooling as the readout. 

\subsection*{Contrastive Learning}

Contrastive learning \cite{he2020momentum} aims at learning representation through contrasting positive data pairs against negative pairs. SimCLR \cite{chen2020simple} demonstrates that contrastive learning for images can greatly benefits from the composition of data augmentations and large batch sizes. Based on InfoNCE loss \cite{oord2018infonce}, SimCLR proposes the NT-Xent loss as given in Equation~\ref{eq:ntxent}:
\begin{equation}
    \mathcal{L}_{i,j} = - \log \frac{\exp(\text{sim}(\pmb{z}_i, \pmb z_j) / \tau)}{\sum_{k=1}^{2N} \mathbbm{1}\{k \neq i\} \exp(\text{sim}(\pmb{z}_i, \pmb z_k) / \tau)},
\label{eq:ntxent}
\end{equation}
where $\pmb z_i$ and $\pmb z_j$ are latent vectors extracted from a positive data pair, $N$ is the batch size, sim$(\cdot)$ measures the similarity between the two vectors, and $\tau$ is the temperature parameter. In our MolCLR, we follow the NT-Xent loss to conduct pre-training on GNN encoders and implement cosine similarity as $\text{sim}(z_i,z_j)=\frac{z_i^T z_j}{\|z_i\|_2 \|z_j\|_2}$. Further investigation of $\tau$ on MolCLR pre-training is included in Supplementary Table 5. Though contrastive learning frameworks have been implemented to various domains, including unstructured graphs \cite{you2020graph}, sentence embeddings \cite{gao2021simcse}, and robotics planning \cite{wang2020cloud}. Contrastive learning has not yet been investigated comprehensively and elaborately for molecule graphs. 

\subsection*{Datasets}

\paragraph{Pre-training Dataset.} For MolCLR pre-training, we use $\sim$10 million unique unlabeled molecule SMILES collected by ChemBERTa \cite{chithrananda2020chemberta} from PubChem \cite{kim2019pubchem}. RDKit \cite{greg2006rdkit} is then utilized to build the molecule graphs and extract chemical features from the SMILES strings. Within the molecule graph, each node represents an atom and each edge represents a chemical bond. We randomly split the pre-training dataset into training and validation set with a ratio of 95/5.

\paragraph{Downstream Datasets.} To benchmark the performance of our MolCLR framework, we use 13 datasets from MoleculeNet \cite{wu2018moleculenet}, containing 44 binary classification tasks and 24 regression tasks in total. These tasks cover molecule properties of multiple domains. For all datasets except QM9, we use the scaffold split to create an 80/10/10 train/valid/test split as suggested in \cite{Hu2020Strategies}. Unlike the common random split, the scaffold split, which is based on molecular substructures, makes the prediction task more challenging yet realistic. QM9 follows the random splitting setting as implementations of most related works \cite{schutt2018schnet, lu2019molecular, Liu2019NGramGS} for comparison. 

\subsection*{Training Details}

Each atom on the molecule graph is embedded by its atomic number and chirality type, while each bond is embedded by its type and direction. 
We implement a 5-layer graph convolutions \cite{kipf2016semi, xu2018how} with ReLU activation as the GNN backbone, and follow the modification in Hu et al. \cite{Hu2020Strategies} to make aggregations compatible with edge features. An average pooling is applied on each graph as the readout operation to extract the 512-dimension molecular representation. An MLP with one hidden layer maps the representation into a 256-dimension latent space. Adam \cite{kingma2014adam} optimizer with weight decay $10^{-5}$ is used to optimize the NT-Xent loss. After the initial 10 epochs with a learning rate, $5\times10^{-4}$, a cosine learning decay is implemented. The model is trained with batch size 512 for the total 50 epochs. 

For the downstream task fine-tuning, we add a randomly initialized MLP on top of the base GNN feature extractor. Softmax cross-entropy loss and $\ell_1$ loss are implemented for classification and regression tasks, respectively. On each task, we conduct 100-epoch fine-tuning of the pre-trained model three times to get the average and standard deviation of performance on the test set. We train the model on training set only and perform search of hyper-parameters on the validation set for the best results. The whole framework is implemented based on Pytorch Geometric \cite{Fey/Lenssen/2019}. More fine-tuning details are included in Supplementary Table 6. 

\subsection*{Baselines}

\paragraph{Supervised learning models.} We comprehensively evaluate the performance of our MolCLR model in comparison with supervised learning methods. For shallow machine learning models, Random Forest (RF) \cite{ho1995random} and Support Vector Machine (SVM) \cite{cortes1995support} are implemented, which take molecular FPs as the input. Multiple GNNs are also included. GCN \cite{kipf2016semi} and GIN \cite{xu2018how, Hu2020Strategies} with edge feature involved in aggregation are considered. Besides, several GNN models which achieve SOTA on several molecular benchmarks are implemented as baselines, i.e., SchNet \cite{schutt2018schnet}, MGCN \cite{lu2019molecular}, and D-MPNN \cite{yang2019analyzing}. These GNNs are designed specifically for molecular. For example, SchNet and MGCN explicitly model quantum interactions within molecules.  

\paragraph{Self-supervised learning models.} To better demonstrate the effectiveness of MolCLR framework, we further include other pre-training or self-supervised learning models as baselines. Hu et al. \cite{Hu2020Strategies} proposes both node-level and graph-level pre-training for molecule graphs. It should be pointed out that though node-level pre-training is based on self-supervision, while the graph-level pre-training is supervised on some molecule property labels \cite{Hu2020Strategies}. N-Gram graph \cite{Liu2019NGramGS} is also implemented, which computes a compact representation directly through the molecule graph.

\section*{Data availability}

The pre-training data and molecular prooerty prediction benchmarks used in this work are available in the both the CodeOcean capsule: \href{https://doi.org/10.24433/CO.8582800.v1}{https://doi.org/10.24433/CO.8582800.v1} \cite{wang2021molclrcode} and the Github repository: \href{https://github.com/yuyangw/MolCLR}{https://github.com/yuyangw/MolCLR}. 

\section*{Code availability}

The code accompanying this work are available in both the CodeOcean capsule: \href{https://doi.org/10.24433/CO.8582800.v1}{https://doi.org/10.24433/CO.8582800.v1} \cite{wang2021molclrcode} and the Github repository: \href{https://github.com/yuyangw/MolCLR}{https://github.com/yuyangw/MolCLR}. 







\newpage

\renewcommand\thefigure{\thesection} 
\renewcommand\thetable{\thesection}

{\parindent0pt 
\Large \textsf{\bf Supplementary Information}
}

\section{Details of Molecular Datasets}

Table~\ref{tb:benchmarks} summarizes all the benchmarks used in our work. These benchmarks from MoleculeNet \cite{wu2018moleculenet} cover a wide variety of molecular properties, including physiology (i.e., BBBP, Tox21, SIDER, ClinTox), biophysics (i.e., BACE, MUV, HIV), physical chemistry (i.e., FreeSolv, Lipo, ESOL), and quantum mechanics (i.e., QM7, QM8, QM9). Also, numbers of data vary significantly among the benchmarks, ranging from less than 1K to more than 130K. All benchmarks except QM9 are scaffold split to train/validation/test sets by the ratio of 8/1/1, which provides a more challenging yet realistic setting. Random splitting is implemented on QM9 following the settings in most related works \cite{schutt2018schnet, lu2019molecular, Liu2019NGramGS} for comparison. ROC-AUC is used as the metric for classification tasks while RMSE and MAE are used for regression tasks. 

\begin{table}[htb!]
  \centering
  \begin{tabular}{llllll}
    \toprule
    Dataset & \# Molecules & \# Tasks & Task type & Metric & Split \\
    \midrule
    BBBP & 2039 & 1 & Classification & ROC-AUC & Scaffold \\
    Tox21 & 7831 & 12 & Classification & ROC-AUC & Scaffold \\
    ClinTox & 1478 & 2 & Classification & ROC-AUC & Scaffold \\
    HIV & 41127 & 1 & Classification & ROC-AUC & Scaffold \\
    BACE & 1513 & 1 & Classification & ROC-AUC & Scaffold \\
    SIDER & 1427 & 27 & Classification & ROC-AUC & Scaffold \\
    MUV & 93087 & 17 & Classification & ROC-AUC & Scaffold \\
    \midrule
    FreeSolv & 642 & 1 & Regression & RMSE & Scaffold \\
    ESOL & 1128 & 1 & Regression & RMSE & Scaffold \\
    Lipo & 4200 & 1 & Regression & RMSE & Scaffold \\
    QM7 & 6830 & 1 & Regression & MAE & Scaffold \\
    QM8 & 21786 & 12 & Regression & MAE & Scaffold \\
    QM9 & 130829 & 8 & Regression & MAE & Random \\
    \bottomrule
  \end{tabular}
  \caption{Summary of all the benchmarks for molecular property predictions used in this work.}
  \label{tb:benchmarks}
\end{table}

We follow Hu et al. \cite{Hu2020Strategies} to build a simple yet unambiguous set of node and bond features to embed the two-dimensional (2D) molecular graph. RDKit is used to convert SMILES to a 2D graph and extract the features. The details of node and edge features can be found in Table~\ref{tb:features}. When a node is masked, the atomic number is set to 119 and chirality to unspecified. 

\begin{table}[htb!]
  \centering
  \begin{tabular}{lll}
    \toprule
    Feature type & Feature name & Range \\
    \midrule
    \multirow{2}{*}{Node feature} & Atomic number & [1, 119] \\
    & Chirality & \{unspecified, tetrahedral CW, tetrahedral CCW, other\} \\
    \midrule
    \multirow{2}{*}{Edge feature} & Bond type & \{single, double, triple, aromatic\} \\
    & Bond direction & \{none, end-upright, end-downright\} \\
    \bottomrule
  \end{tabular}
  \caption{Node and edge features used in MolCLR.}
  \label{tb:features}
\end{table}

\newpage

\section{Detailed Results of QM9}

Table~\ref{tb:qm9} reports detailed results on QM9 database. The property name, unit, mean and std of test MAE for all the models are included. Not surprisingly, SchNet \cite{schutt2018schnet} and MGCN \cite{lu2019molecular} outperform the other models greatly. These two models successfully develop interaction layers, which elaborately take quantum interactions into consideration as titles of both works indicate. Besides, both models include 3D positional information as the input, which benefits quantum mechanics property predictions. However, MolCLR pre-training is still demonstrated to be effective on this challenging benchmark. MolCLR shows better prediction accuracy in 7 out 8 tasks among all the pre-training/self-supervised models. MolCLR\textsubscript{GIN} surpasses Hu et al. \cite{Hu2020Strategies} in all the tasks, which also utilizes GIN as the encoder. Besides, in comparison to GCN and GIN trained via supervised learning, MolCLR\textsubscript{GCN} and MolCLR\textsubscript{GIN} improve the performance on all the tasks within QM9. MolCLR also obtains lower test MAE when set side-by-side with another supervised baseline, D-MPNN \cite{yang2019analyzing}.

\begin{table}[htb!]
  \centering
  \resizebox{\textwidth}{!}{\begin{tabular}{lllllllll}
    \toprule
    Property & $\epsilon_{HOMO}$ & $\epsilon_{LUMO}$ & $\Delta\epsilon$ & ZPVE & $\mu$ & $\alpha$ & $\langle R^{2}\rangle$ & $C_v$\\
    Unit & eV & eV & eV & eV & D & bohr$^3$ & bohr$^2$ & cal/mol K\\
    \midrule
    RF & 0.186$\pm$0.001 & 0.276$\pm$0.002 & 0.269$\pm$0.001 & 0.276$\pm$0.000 & 0.658$\pm$0.004 & 3.245$\pm$0.015 & 121.837$\pm$0.124 & 1.738$\pm$0.003 \\
    SVM & 0.148$\pm$0.000 & 0.234$\pm$0.002 & 0.248$\pm$0.004 & 0.157$\pm$0.000 & 0.750$\pm$0.004 & 4.065$\pm$0.057 & 189.510$\pm$1.078 & 1.795$\pm$0.010 \\
    GCN \cite{kipf2016semi} & 0.115$\pm$0.010 & 0.133$\pm$0.007 & 0.174$\pm$0.013 & 0.075$\pm$0.018 & 0.532$\pm$0.015 & 1.495$\pm$0.338 & 43.325$\pm$15.140 & 0.514$\pm$0.209 \\
    GIN \cite{xu2018how} & 0.097$\pm$0.005 & 0.103$\pm$0.010 & 0.138$\pm$0.004 & 0.055$\pm$0.021 & 0.483$\pm$0.004 & 1.315$\pm$0.405 & 35.278$\pm$6.779 & 0.457$\pm$0.073 \\
    SchNet \cite{schutt2018schnet} & 0.041$\pm$0.001 & 0.034$\pm$0.003 & 0.063$\pm$0.002 & 0.002$\pm$0.000 & 0.033$\pm$0.001 & 0.235$\pm$0.061 & 0.073$\pm$0.002 & 0.033$\pm$0.000  \\
    MGCN \cite{lu2019molecular} & 0.042$\pm$0.001 & 0.057$\pm$0.002 & 0.064$\pm$0.001 & 0.001$\pm$0.000 & 0.056$\pm$0.002 & 0.030$\pm$0.007 & 0.113$\pm$0.001 & 0.038$\pm$0.001 \\
    D-MPNN \cite{yang2019analyzing} & 0.093$\pm$0.005 & 0.106$\pm$0.002 & 0.148$\pm$0.003 & 0.037$\pm$0.004 & 0.450$\pm$0.006 & 0.493$\pm$0.008 & 24.371$\pm$0.922 & 0.244$\pm$0.005 \\
    \midrule
    HU. et.al \cite{Hu2020Strategies} & 0.116$\pm$0.000 & 0.118$\pm$0.000 & 0.161$\pm$0.001 & 0.083$\pm$0.001 & 0.543$\pm$0.001 & 1.725$\pm$0.008 & 55.418$\pm$0.291 & 0.705$\pm$0.012 \\
    N-Gram \cite{Liu2019NGramGS} & 0.142$\pm$0.001 & 0.138$\pm$0.001 & 0.193$\pm$0.001 & 0.009$\pm$0.000 & 0.540$\pm$0.002 & 0.611$\pm$0.022 & 59.137$\pm$0.178 & 0.334$\pm$0.007 \\
    MolCLR\textsubscript{GCN} & 0.104$\pm$0.000 & 0.110$\pm$0.001 & 0.149$\pm$0.001 & 0.045$\pm$0.004 & 0.507$\pm$0.002 & 0.644$\pm$0.053 & 26.600$\pm$0.257 & 0.259$\pm$0.011 \\
    MolCLR\textsubscript{GIN} & 0.087$\pm$0.000 & 0.092$\pm$0.000 & 0.127$\pm$0.000 & 0.033$\pm$0.004 & 0.464$\pm$0.001 & 0.463$\pm$0.017 & 17.425$\pm$0.919 & 0.164$\pm$0.002 \\
    \bottomrule
  \end{tabular}}
  \caption{Test MAE of different models for each property in QM9.}
  \label{tb:qm9}
\end{table}

\newpage

\section{Investigation of Pre-training Datasets for MolCLR}

MolCLR pre-training makes use of the large unlabeled molecular data. We also investigate whether pre-training on certain dataset benefits molecular property predictions of its own. To this end, we conduct MolCLR pre-training on MUV and QM9 as shown in Table~\ref{tb:pretrain_dataset}, since these are the two largest datasets in MoleculeNet \cite{wu2018moleculenet}. Within the table, MolCLR\textsubscript{PubChem} denotes MolCLR framework pre-trained on the $\sim$10M unlabeled molecules from PubChem \cite{kim2019pubchem}. MolCLR\textsubscript{MUV} and MolCLR\textsubscript{QM9} indicates pre-training on MUV and QM9, respectively. The training and fine-tuning follow the same setting reported in the main manuscript. To avoid data leakage, we split the MUV and QM9 into train/validation/test by the ratio of 8:1:1 and only pre-train the models on the training splits. When conducting fine-tuning on MUV, MolCLR pre-training on MUV improves the test ROC-AUC by 15.4\% and MolCLR pre-trained on QM9 also obtains a great improvement by 14.9\% in comparison with no pre-training. Not surprisingly, MolCLR pre-trained on the 10M dataset performs the best better as it benefits from larger unlabeled molecular data. Similarly, on QM9, pre-training on MUV and QM9 decreases the test MAE by 1.553 and 2.010, respectively. As expected, MolCLR\textsubscript{PubChem} achieves the larges improvement by 2.384 on QM9. Therefore, pre-training on the dataset itself via MolCLR boosts the performance significantly. Also, the pre-trained model on one dataset can be directly transferred to another and outperforms training from scratch. 

\begin{table}[htb!]
  \centering
  \begin{tabular}{llllll}
    \toprule
    & Metric & Supervised & MolCLR\textsubscript{MUV} & MolCLR\textsubscript{QM9} & MolCLR\textsubscript{PubChem} \\
    \midrule
    MUV & ROC-AUC (\%) & 71.8$\pm$2.5 & 87.2$\pm$2.1 & 86.7$\pm$2.8 & 88.6$\pm$2.2 \\
    \midrule
    QM9 & MAE & 4.741$\pm$0.912 & 3.188$\pm$0.441 & 2.731$\pm$0.019 & 2.357$\pm$0.118 \\
    \bottomrule
  \end{tabular}
  \caption{Comparison of MolCLR pre-training on different datasets. Test ROC-AUC (\%) are reported for MUV and MAE for QM9. Supervised indicates supervised learning with no pre-training. MolCLR\textsubscript{MUV} MolCLR\textsubscript{QM9}, and  MolCLR\textsubscript{PubChem} denote MolCLR pre-training on the MUV, QM9, and $\sim$10M PubChem, respectively. }
  \label{tb:pretrain_dataset}
\end{table}

\begin{figure}[hbt!]
\centering
    \begin{subfigure}{0.48\textwidth}
      \centering
      \includegraphics[width=\linewidth]{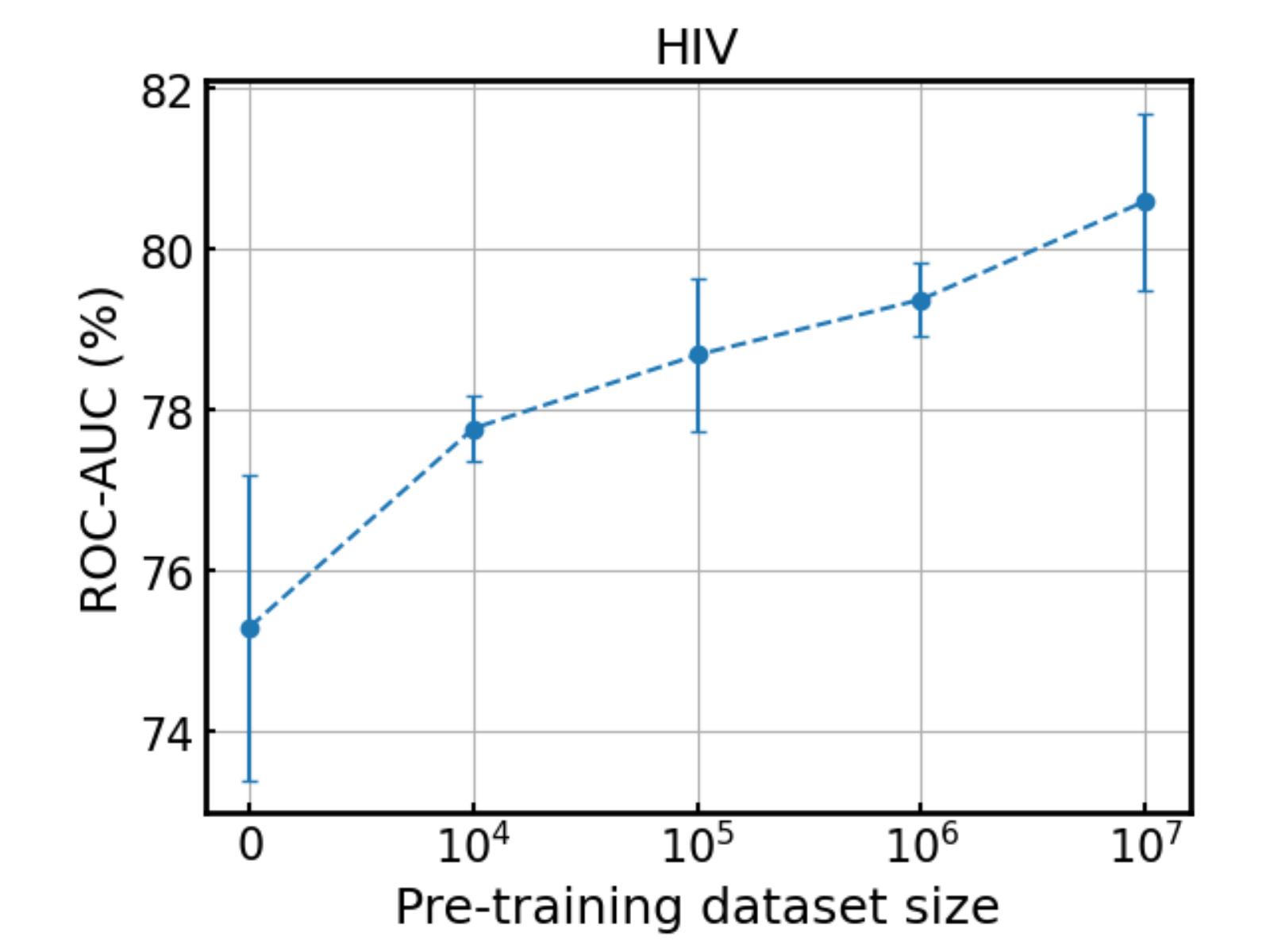}
      \caption{}
      \label{fig:hiv_datasize}
    \end{subfigure}
    \begin{subfigure}{0.48\textwidth}
      \centering
      \includegraphics[width=\linewidth]{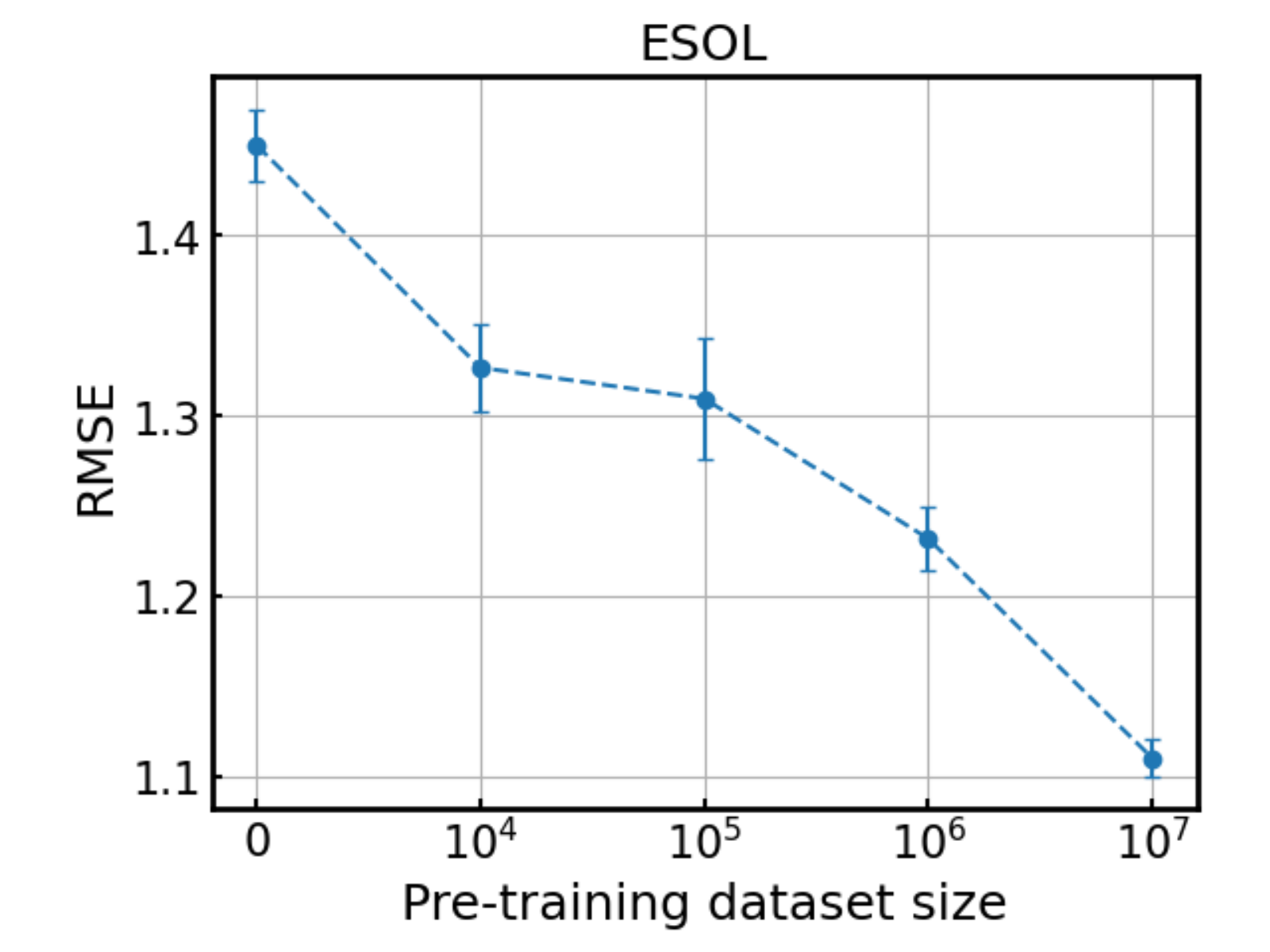}
      \caption{}
      \label{fig:esol_datasize}
    \end{subfigure}
\caption{Results of MolCLR pre-training on different dataset sizes. (a) Test ROC-AUC (\%) on HIV. (b) Test RMSE on ESOL.}
\label{fig:datasize}
\end{figure}

We further probe the influence of the magnitude of pre-training datasets on MolCLR. Subsets of size 10K, 100K, and 1M are randomly sampled from the whole $\sim$10M PubChem pre-training dataset. Figure~\ref{fig:datasize}(a) and Figure~\ref{fig:datasize}(b) report the test results of different pre-training data size on HIV and ESOL databases. Pre-training dataset size 0 indicates supervised learning is directly conducted without pre-training. As the number of data increases, the averaged test HIV ROC-AUC increases from 75.3 to 80.6. Similarly, the larger the dataset, the lower test RMSE on ESOL is observed. Also, even pre-training on a small dataset, i.e., 10K molecules, GNN models gain obvious improvements in comparison to supervised learning. For example, pre-training on 10K data improves ROC-AUC by 2.5\% on HIV and decreases RMSE by 0.12 on ESOL. It is demonstrated that MolCLR benefits from the large dataset, and therefore can be widely used for the huge unlabeled molecule data. On the other hand, MolCLR pre-training on a small dataset still boosts the performance compared to supervised learning, which demonstrates the effectiveness of the contrastive learning framework on molecule graphs. 

\newpage

\section{Visualization of MolCLR Representations}

\begin{figure}[hbt!]
\centering
    \begin{subfigure}{0.49\textwidth}
      \centering
      \includegraphics[width=\linewidth]{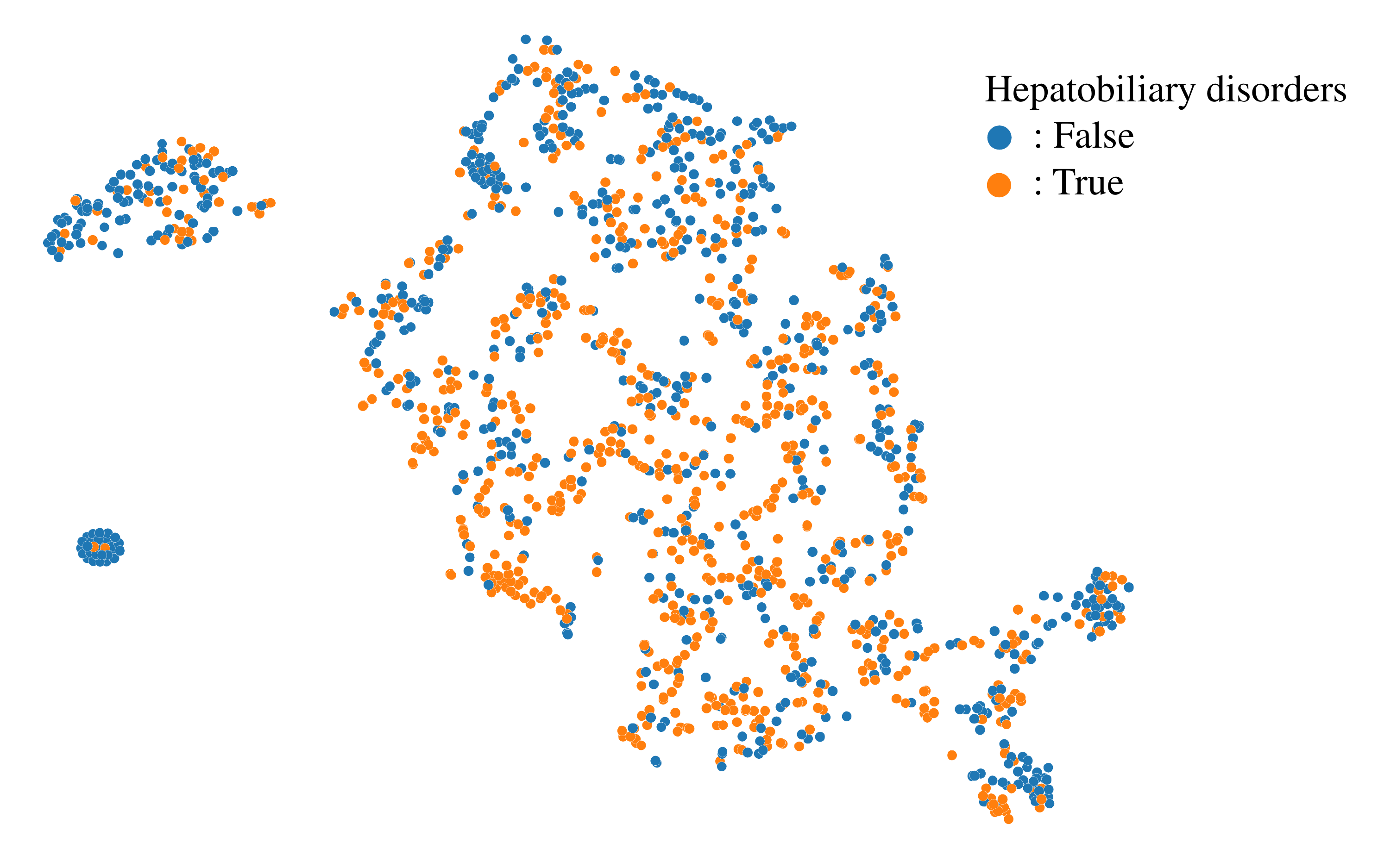}
      \caption{}
      \label{fig:sider}
    \end{subfigure}
    \begin{subfigure}{0.49\textwidth}
      \centering
      \includegraphics[width=\linewidth]{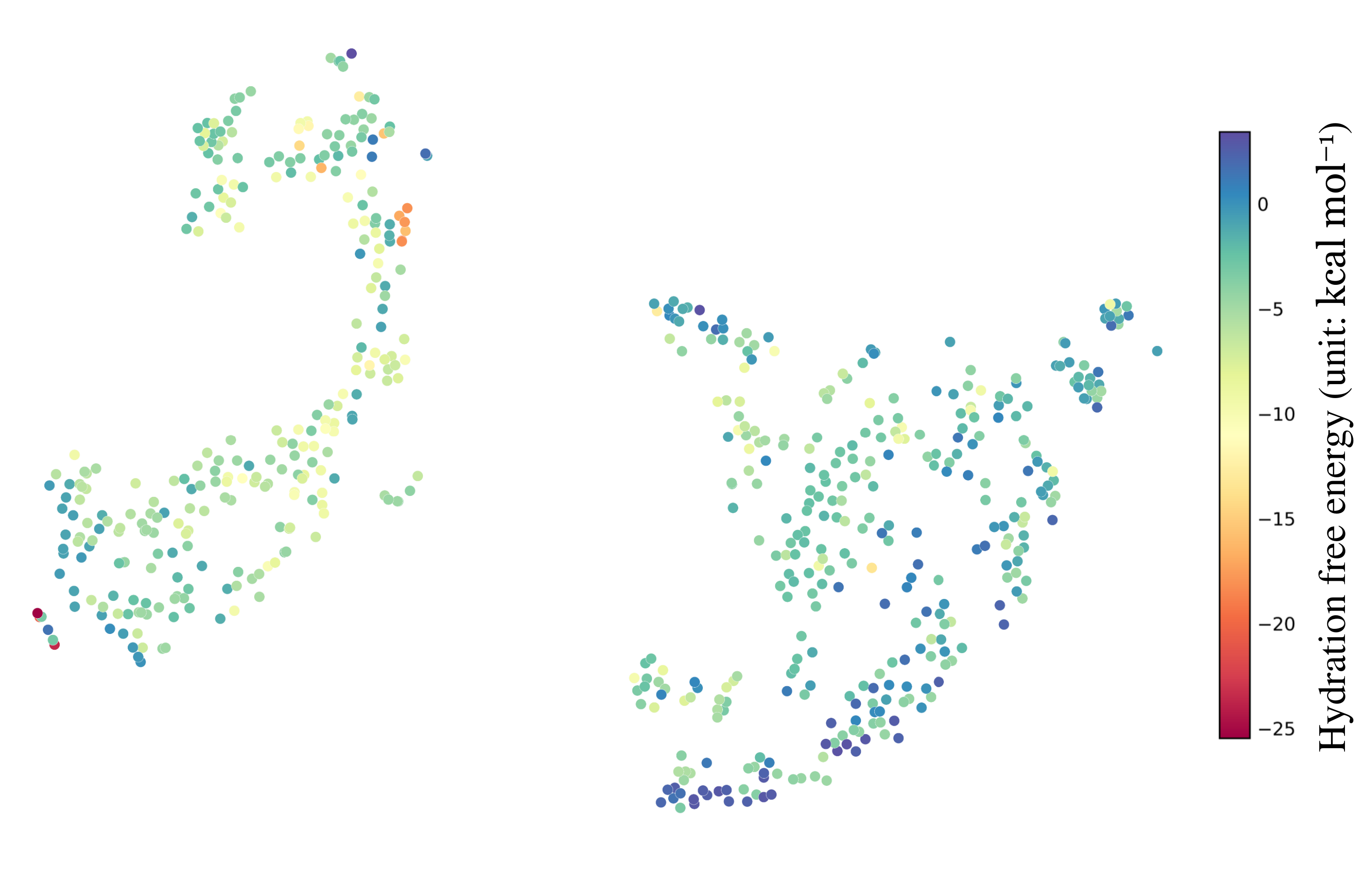}
      \caption{}
      \label{fig:freesolv}
    \end{subfigure}
    \begin{subfigure}{0.49\textwidth}
      \centering
      \includegraphics[width=\linewidth]{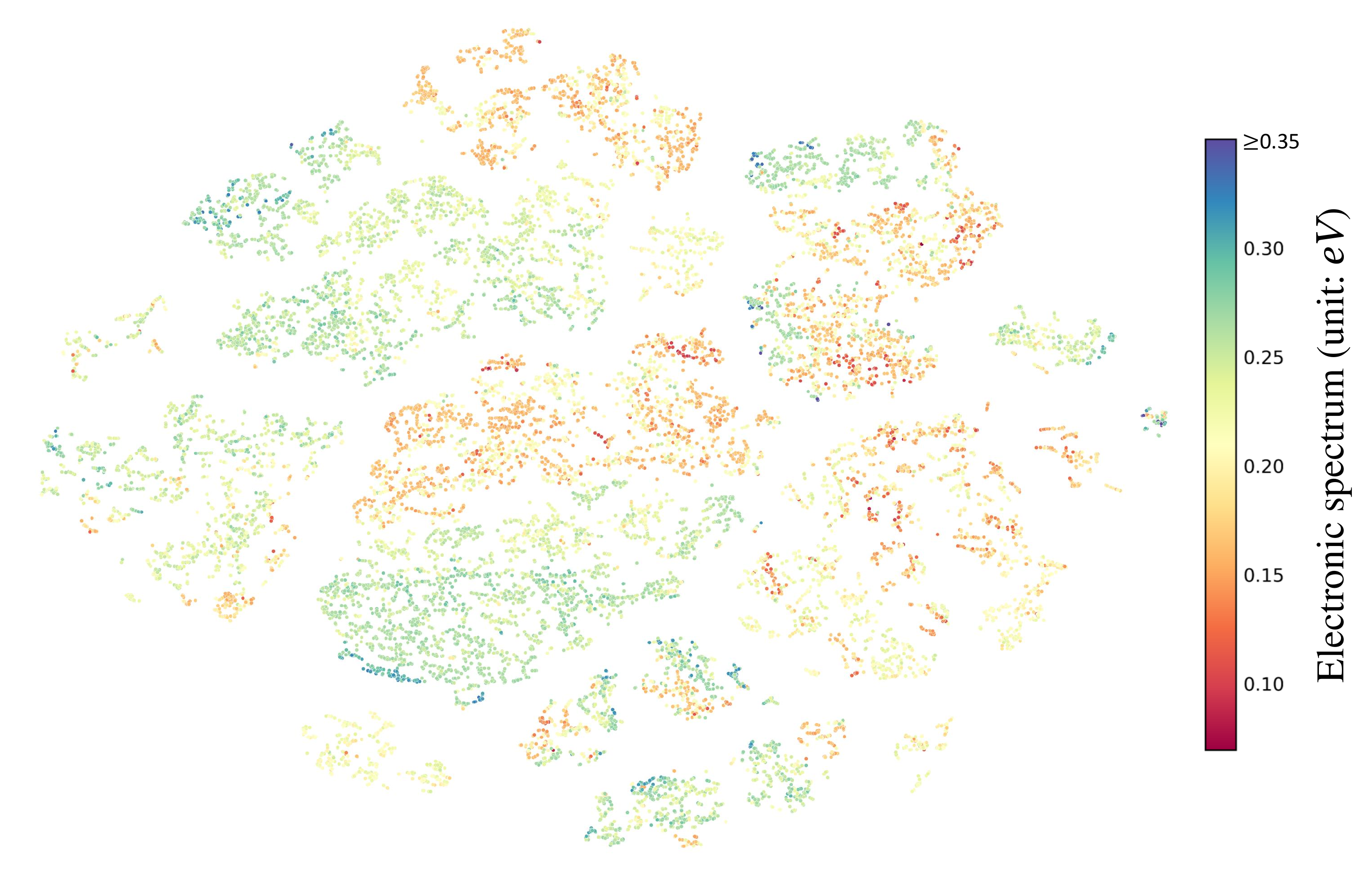}
      \caption{}
      \label{fig:qm8}
    \end{subfigure}
    \begin{subfigure}{0.49\textwidth}
      \centering
      \includegraphics[width=\linewidth]{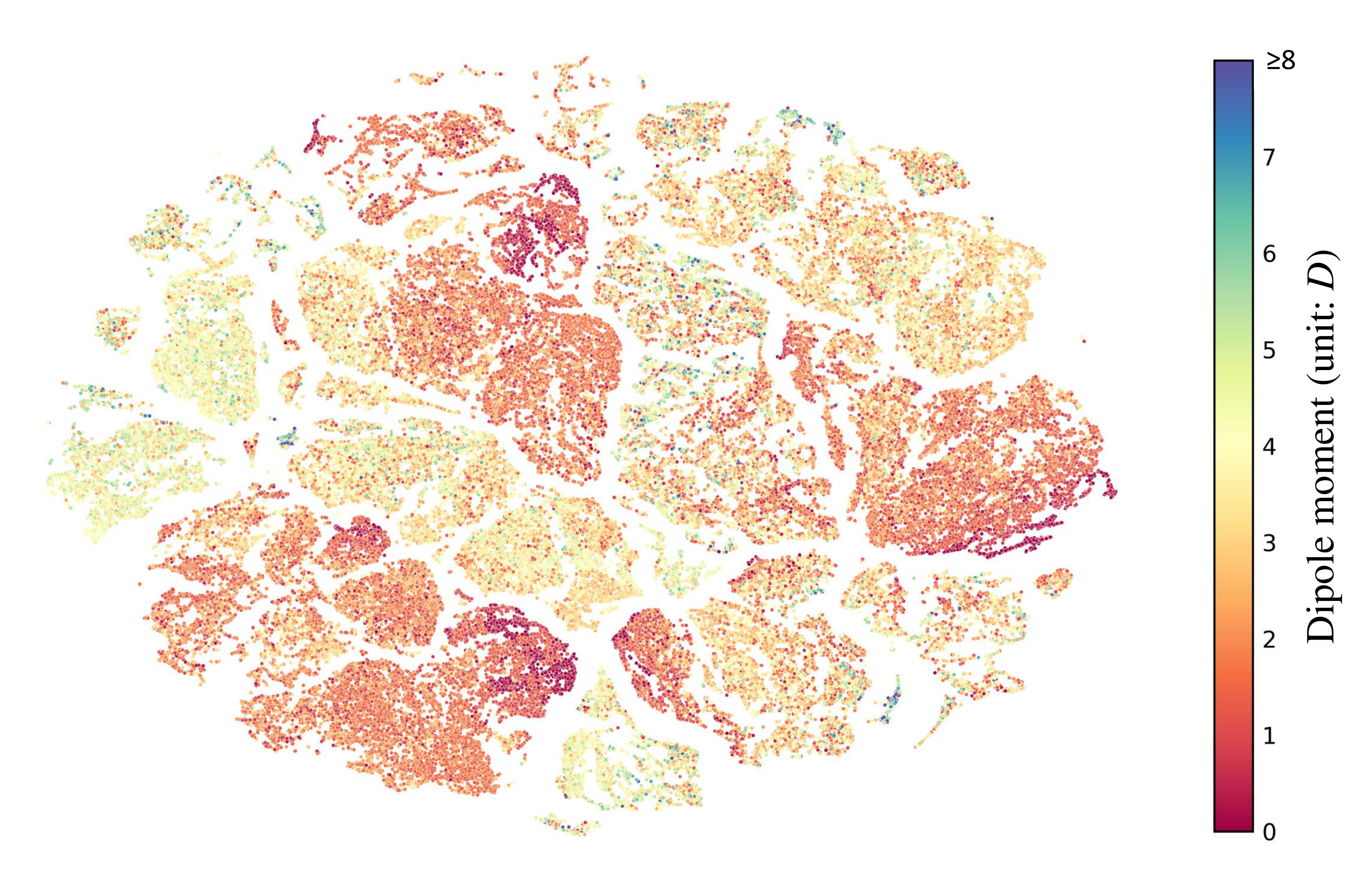}
      \caption{}
      \label{fig:qm9}
    \end{subfigure}
\caption{Two-dimensional t-SNE embedding of the molecular representations learned by our MolCLR pre-training. (a) Molecules from SIDER database and color indicates whether the molecule causes hepatobiliary disorder side effect. (b) Molecules from FreeSolv database and color indicates hydration free energy of each molecule. (c) Molecules from QM8 database and color indicates the electronic spectrum calculated from CC2 of each molecule. (d) Molecules from QM9 database and color indicates the averaged electronic spectrum $\mu$ of each molecule.}
\label{fig:tsne_label}
\end{figure}

Besides, to illustrate the representations from the pre-trained MolCLR, we visualize the molecule features via t-SNE, where molecules are from various databases and colored by corresponding property labels (Figure~\ref{fig:tsne_label}). Notice that all the features are extracted directly from pre-trained MolCLR without fine-tuning. Namely, the model has no access to the molecular property labels during training. Figure~\ref{fig:tsne_label} shows molecules from SIDER \cite{kuhn2016sider}, FreeSolv \cite{mobley2014freesolv}, QM8 \cite{ruddigkeit2012enumeration, ramakrishnan2015electronic}, QM9 \cite{ramakrishnan2014quantum}. Features from pre-trained MolCLR show clustering based on the labels, even without accessing labels during training. For instance, in Figure~\ref{fig:tsne_label}(d), molecules are colored by the dipole moment $\mu$. Molecules with relatively high $\mu$ (green and blue) are clustered on the bottom right, whereas molecules with low $\mu$ (dark red) are clustered in the center of the plot. Similar clustering trends can also be observed in other t-SNE visualizations in Figure~\ref{fig:tsne_label}.

\newpage

\section{More Results of Molecule Retrieval via MolCLR}

\begin{figure}[hbt!]
\centering
    \begin{subfigure}{0.8\textwidth}
      \centering
      \includegraphics[width=\linewidth]{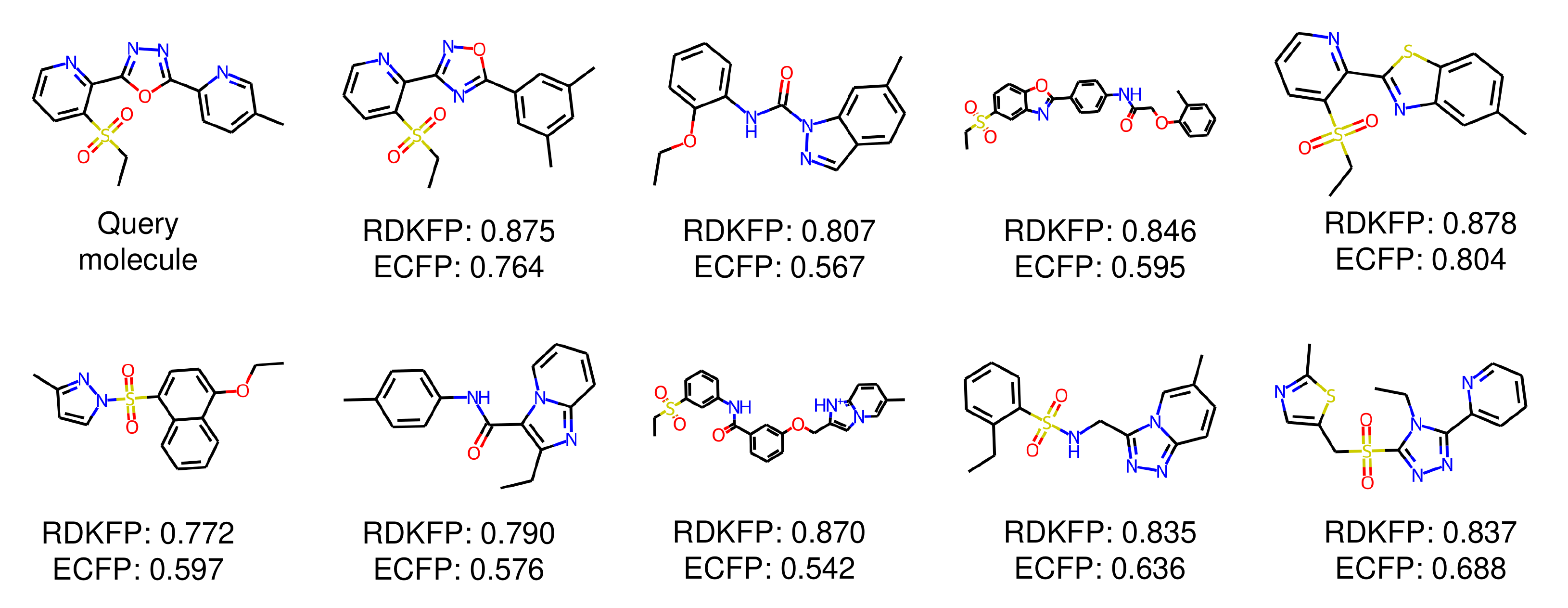}
      \caption{}
    \end{subfigure}
    \begin{subfigure}{0.8\textwidth}
      \centering
      \includegraphics[width=\linewidth]{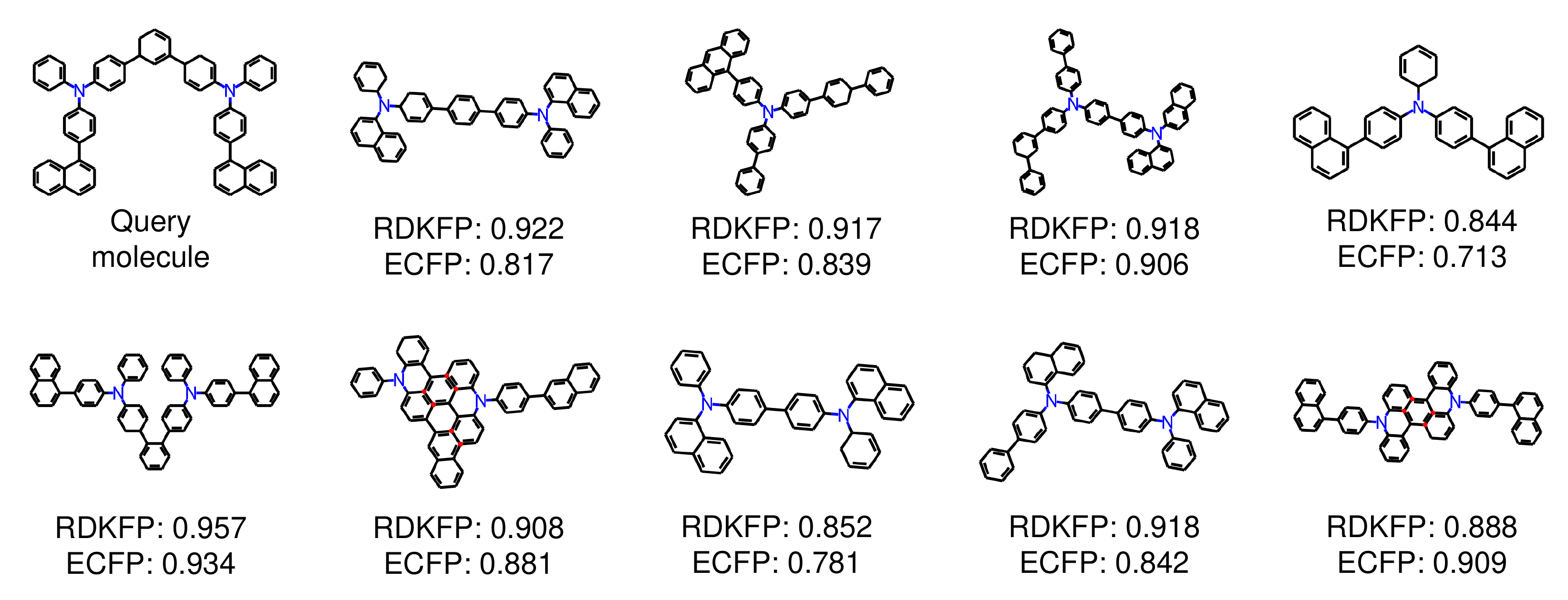}
      \caption{}
    \end{subfigure}
    \begin{subfigure}{0.8\textwidth}
      \centering
      \includegraphics[width=\linewidth]{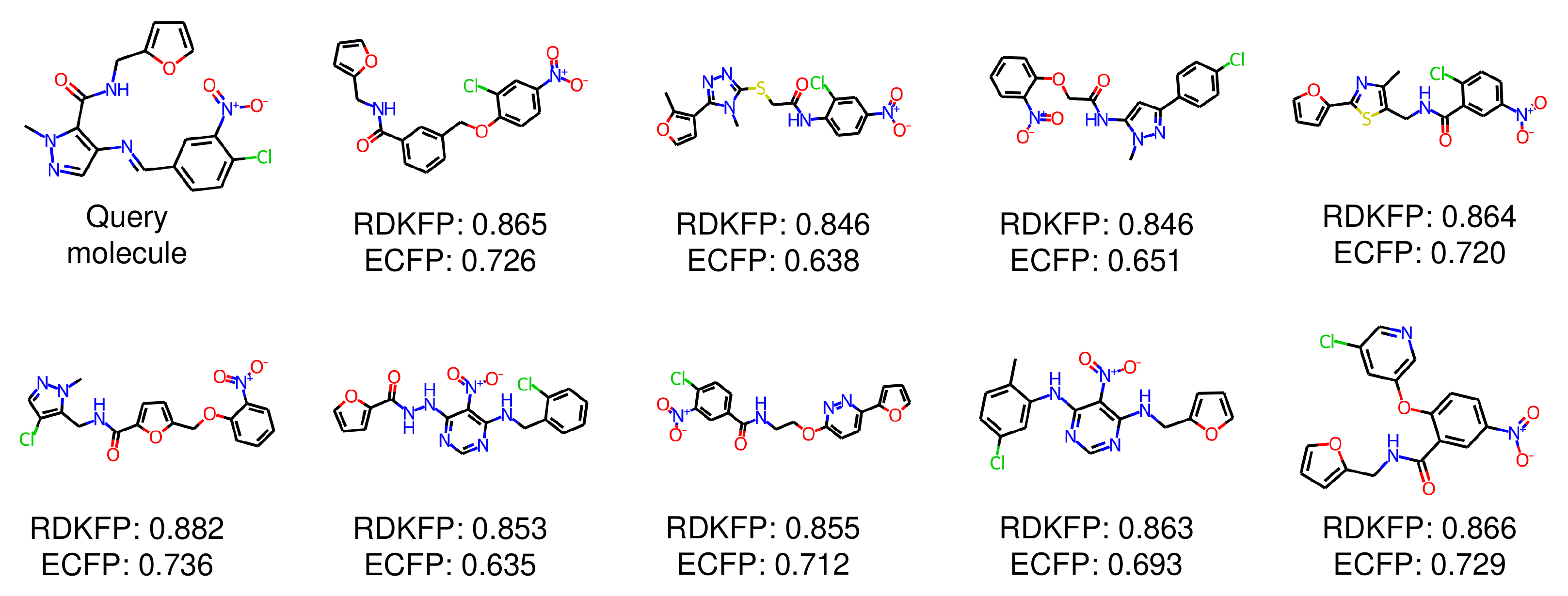}
      \caption{}
    \end{subfigure}
\caption{Three Query molecules (PubChem ID (a) 130187714 (b) 132175476 (c) 4862714) and 9 closest molecules for each query molecule in MolCLR representation domain with RDKFP and ECFP similarities labeled.}
\label{fig:sim_mol_si}
\end{figure}

In this section, more examples of molecule retrieval based on MolCLR-learned representations are shown in Figure~\ref{fig:sim_mol_si}. Nine molecules that are closest to the query molecule in the MolCLR representation domain are listed with RDKFP and ECFP similarities labeled. Notably, molecules with close MolCLR representations also have high FP similarities. Also, the selected molecules share similar structures and functional groups. For instance, in Figure~\ref{fig:sim_mol_si}(a), all listed molecules share functional groups like sulfonyl groups and nitrogen heterocycles. Also, in Figure~\ref{fig:sim_mol_si}(b), the first molecule at the second row is exactly the same as the query molecule except for few carbon-carbon bonds. These examples further demonstrate that through contrastive learning, MolCLR automatically learns chemically meaningful representations.

\newpage

\section{Temperature in Contrastive Loss}

The choice of the temperature parameter $\tau$ in NT-Xent loss \cite{chen2020simple} impacts the performance of contrastive learning \cite{chen2020simple}. An appropriate $\tau$ benefits the model to learn from hard negative samples. To investigate $\tau$ for molecule representation learning, we train MolCLR with three different temperatures: $0.05$, $0.1$, and $0.5$ as shown in Table~\ref{tb:temperature}. We report the averaged ROC-AUC (\%) over all the seven classification benchmarks using 25\% subgraph removal as the augmentation strategy. It is demonstrated that $\tau=0.1$ performs the best in the downstream molecular tasks. Therefore, we use $\tau=0.1$ as the temperature in the following experiments. 

\begin{table}[htb!]
  \centering
  \begin{tabular}{llll}
    \toprule
    Temperature ($\tau$) & 0.05 & 0.1 & 0.5 \\
    \midrule
    ROC-AUC (\%) & 76.8$\pm$1.2 & 80.2$\pm$1.3 & 78.4$\pm$1.7 \\
    \bottomrule
  \end{tabular}
\caption{Influence of temperature $\tau$ in NT-Xent loss for MolCLR. Mean and standard deviation of all the seven classification benchmarks are reported.}
\label{tb:temperature}
\end{table}

\newpage

\section{Fine-tuning Details}

During fine-tuning for each downstream task, we randomly search the hyper-parameters to find the best performing setting on the validation set and report the results on the test set. Table~\ref{tb:finetune} lists the combinations of different hyper-parameters. Besides, we also consider if cosine annealing learning rate decay \cite{loshchilov2016sgdr} improves the fine-tuning performance. In addition, we randomly pick MolCLR-trained GNNs at different epoch as the initialization for fine-tuning. 

\begin{table}[htb!]
  \centering
  \begin{tabular}{llll}
    \toprule
    Name & Description & Range \\
    \midrule
    \texttt{batch\_size} & Input batch size & \{32, 128, 256\} \\ 
    \texttt{lr} & Initial learning rate for MLP head & \{$5 \times 10^{-4}$, $10^{-3}$\} \\
    \texttt{lr\_base} & Initial learning rate for the pre-trained GNN base & \{$5 \times 10^{-5}$, $10^{-4}$, $2 \times 10^{-4}$, $5 \times 10^{-4}$\} \\
    \texttt{dropout} & Dropout ratio for the GNN & \{0, 0.1, 0.3, 0.5\} \\
    \texttt{n\_layer} & Number of hidden layers in MLP & \{1, 2\} \\
    \texttt{hidden\_size} & Size of hidden layers in MLP & \{256\} \\
    \texttt{activation} & Nonlinear activation function in MLP & \{ReLU\cite{maas2013rectifier}, Softplus\cite{glorot2011deep}\} \\
    \bottomrule
  \end{tabular}
\caption{Fine-tuning hyper-parameters for pre-trained MolCLR model.}
\label{tb:finetune}
\end{table}

\end{document}